\documentclass{article} % For LaTeX2e
\usepackage{ragslab}
\usepackage{natbib}
\usepackage{amsmath,amssymb}
\usepackage{booktabs}
\usepackage{amsmath,amsfonts,bm}

\def\eqref#1{equation~\ref{#1}}
\def\1{\bm{1}}

\DeclareMathAlphabet{\mathsfit}{\encodingdefault}{\sfdefault}{m}{sl}
\SetMathAlphabet{\mathsfit}{bold}{\encodingdefault}{\sfdefault}{bx}{n}

\supportlogo[height=0.75cm]{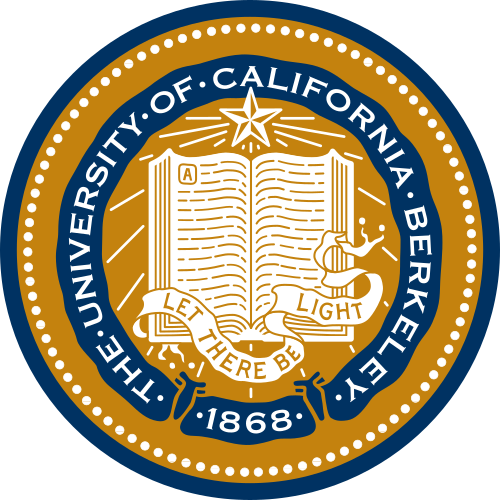}
\supportlogo[height=0.50cm]{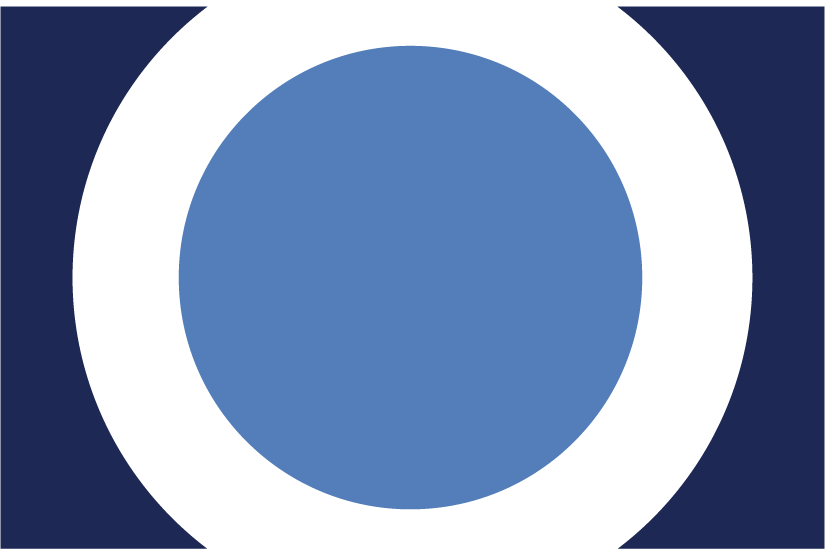}
\lablogo[height=0.75cm]{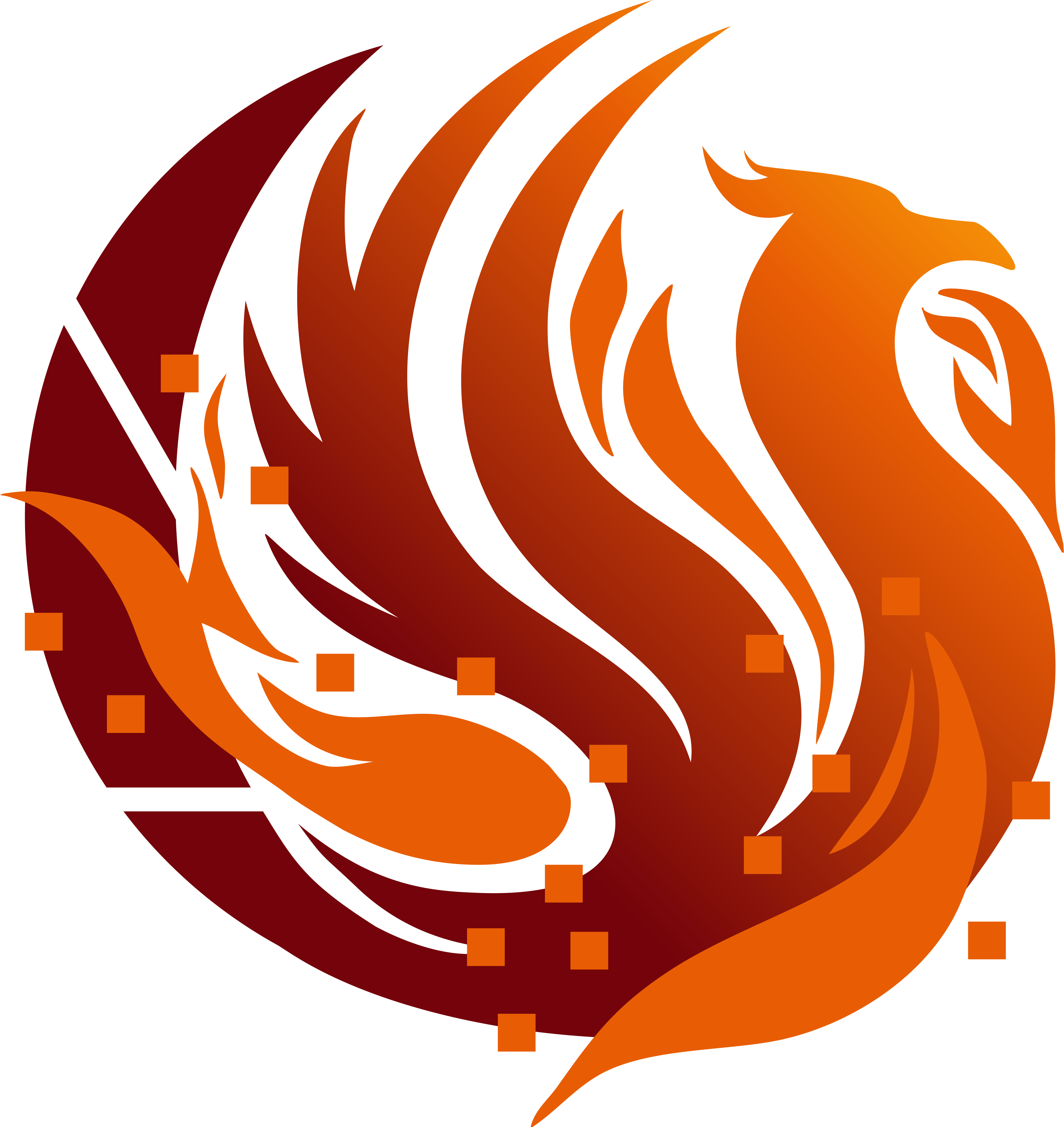}

\title{Learning the Shared Latent Structure of Cryo-EM Inverse Problems}

\author[1,2]{Nithin Chalapathi}
\author[1]{Theo Sternlieb}
\author[1]{Ethan Han}
\author[2]{Sonya M. Hanson}
\author[2]{Pilar Cossio}
\author[1]{Aditi S. Krishnapriyan}
\author[3]{Ritwik Gupta}

\affiliation[1]{University of California, Berkeley}
\affiliation[2]{Flatiron Institute}
\affiliation[3]{University of Maryland, College Park

\date{\today}
\correspondence{Nithin Chalapathi at \email{nithinc@berkeley.edu}}
\code{\url{https://github.com/nithinvc/EMber}}}

\usepackage[utf8]{inputenc} % allow utf-8 input
\usepackage[T1]{fontenc}    % use 8-bit T1 fonts
\usepackage{hyperref}       % hyperlinks
\usepackage{url}            % simple URL typesetting
\usepackage{booktabs}       % professional-quality tables
\usepackage{amsfonts}       % blackboard math symbols
\usepackage{nicefrac}       % compact symbols for 1/2, etc.
\usepackage{microtype}      % microtypography
\usepackage{xcolor}         % colors
\usepackage{amsmath}
\usepackage{graphicx}
\usepackage{wrapfig}         % text flowing beside compact figures and tables
\usepackage{needspace}       % keep wrapped figures clear of the bottom margin
\usepackage{colortbl}       % table cell/row/column coloring
\usepackage{array}          % extended column specs (needed for \columncolor)
\usepackage{capt-of} % TODO: double check this is ok
\usepackage{newtxtext}

\usepackage{silence}
\colorlet{header}{paper-deep}   % subtle warm tint for headers

\newcommand{\Fref}[1]{Fig.~\ref{#1}}
\newcommand{\model}{{{\emph{EMber}}}}

\abstract{Understanding biomolecular function requires characterizing three-dimensional structure and conformational variability, for which cryogenic electron microscopy (cryo-EM) is a powerful tool.
Cryo-EM produces noisy two-dimensional particle images in which molecular structure, particle pose, and imaging conditions are jointly encoded.
Recovering biological information from these observations spans inverse problems including pose estimation, heterogeneity analysis, and reconstruction.
These tasks are usually addressed independently despite shared protein structure and image formation, motivating a common representation across inverse problems.
Since biologically meaningful differences may be spatially localized, we study self-distillation with complementary global and masked patch objectives.
Because experimental dataset discrimination can reflect acquisition-specific cues, we develop controlled synthetic and experimental evaluations and find that structural discrimination benefits from dense spatial features.
We introduce the~\model~family of vision transformers, pre-trained on \textgreater50 million experimental particle images from 214 EMPIAR datasets together with simulated projections from 268,000 molecular assemblies.
\model~improves upon existing methods at conformational and compositional discrimination by 14.85 and 14.26 percentage points, estimates conformational populations to within 1.59pp error with a linear probe, and produces particle assignments yielding reference-quality reconstructions in fewer epochs.
We demonstrate how to learn representations that capture biological information for use across multiple cryo-EM inverse problems.
}

\begin{document}

\maketitle

\section{Introduction}
\label{ref:introduction}

The functions of proteins are closely tied to their three-dimensional structure. Proteins, however, are dynamic objects. They occupy ensembles of conformations that result in markedly different function.
Characterizing these structural ensembles is a central problem in structural biology.

Cryogenic electron microscopy (cryo-EM) enables the study of these ensembles by imaging large numbers of individual protein particles, each producing a two-dimensional projection of a three-dimensional molecular structure with noise levels one to three orders of magnitude larger than those encountered in natural images.
Each particle image jointly reflects the underlying molecular conformation, particle pose, and acquisition-dependent imaging conditions.
As such, recovering structural information requires reasoning from weak observations where the sources of variation are unknown and entangled.

Cryo-EM workflows address this ambiguity through a multitude of computational tasks, including particle classification, pose estimation, heterogeneity analysis, and 3D reconstruction.
While these tasks have different objectives, they repeatedly recover and use overlapping structural and imaging information from the same particle images.
Nevertheless, computational methods are largely applied on a per-experiment basis, while learned models are typically developed for a particular task or molecular system.
Because proteins share common structural features, training a distinct model for each system can be redundant.
This raises a broader question: \textbf{can general protein representations be learned from diverse cryo-EM data and reused across proteins and downstream tasks?}\newline

This question encapsulates an under-specified representation learning problem.
Ground truth labels describing the molecular state of individual particles are rarely available.
At the same time, not all variation between particle images is equally useful.
For example, acquisition-specific signals such as sample preparation procedures can distinguish datasets without reflecting differences in molecular structure, while biologically meaningful conformational changes may affect only a small region of a projected molecule.
A useful general protein representation must preserve molecular information in the presence of strong imaging variation while retaining spatially localized features that can capture subtle structural differences.

The scarcity of structural labels at a particle level makes self-supervised learning a natural learning setting. However, different self-supervised objectives, e.g. contrastive, reconstruction-based, or self-distillation, impose different biases on the information is preserved or discarded in a representation. We consider which learning objective is better suited to the qualities of cryo-EM data and the goal of learning generalizable protein representations.
% Contrastive objectives encourage agreement across paired image-label observations but are unable to account for variation in the forward model.
% Reconstruction-based objectives require recovering image content, including variation that may be dominated by noise or acquisition effects.
% We investigate self-distillation with complementary global and masked-patch objectives, which directly learns both image-level and spatially resolved features.

% Keeping in mind that we have to account for the forward model and noise model, while preserving the structural information needed, we discuss the form / style / characteristics of a self-supervised setting that can retain this information.

We train the~\model~family of vision transformers, ranging from 21M to 1.1B parameters, on \textgreater50 million experimental particles from 214 EMPIAR datasets together with simulated projections from 268,000 molecular assemblies.
We construct self-supervised views directly through the cryo-EM forward model and combine exact structural correspondence from simulation with the diversity of experimental data (Sections~\ref{sec:methods} and~\ref{sec:data-view-construction}).
Across controlled synthetic and experimental evaluations,~\model~improves conformational and compositional discrimination by up to 14 percentage points while showing that dense spatial features retain structural information lost by global pooling (Sections~\ref{sec:cryoppp-knn}).
Frozen~\model~representations further estimate conformational populations to within 1.59 percentage points using only a linear probe (Section~\ref{sec:spike-pop-est}), and can be inserted into a reconstruction pipeline to produce particle groupings yielding near-reference reconstructions after two pose-search epochs (Section~\ref{sec:synthetic-reconstruction}).
Together, these results show that a single pre-trained particle representation can be reused across several cryo-EM inverse problems.

% NOTE: merged the problem-setup with the image formation text
% \input{problem-setup}
\section{Cryo-EM as a Shared Inverse Problem}
\label{sec:cryo-em-background}

\begin{figure}[htbp!]
    \centering
    \includegraphics[width=1.0\linewidth]{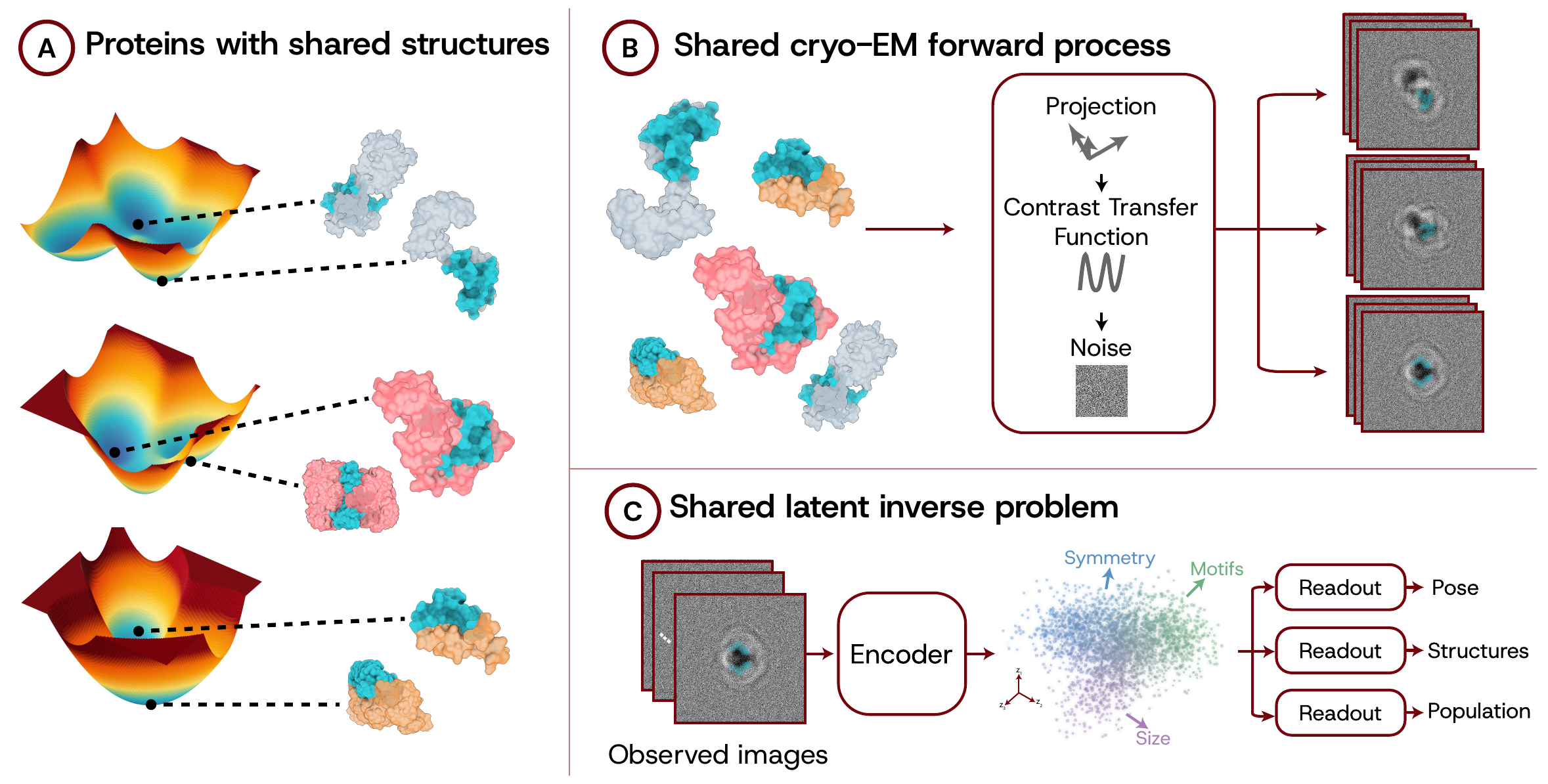}
    \caption{\textbf{Cryo-EM overview and shared inverse problems}. 
    (A) Proteins share similar structural features across molecular systems.
    (B) Cryo-EM images are generated with an image formation process involving projection, imaging effects, and measurement noise.
    (C) Downstream tasks attempt to recover different variables of the same underlying generative process, relying on overlapping structural and imaging information.}
    \label{fig:overview}
\end{figure}

A cryo-EM experiment forms noisy, two-dimensional projections of three-dimensional molecular structures. Let $X_k \in \mathbb{R}^{N\times3}$ be a molecular state with population distribution $\pi(X_k)$. $\rho(X_k): \mathbb{R}^3 \rightarrow \mathbb{R}$ is the corresponding electron density. For particle $i$, the image formation process can abstractly be written as
\begin{align}
    \label{eq:image-formation}
    % y_i = H_i \star \int X_k(R_i ^T r) dz + \epsilon_i
    %y_i = \mathrm{PSF}_i \star \int R_i \cdot \rho(X_k) dz + \epsilon_i
    y_i = \mathrm{PSF}_i \star \mathcal{T}_{t_i}\mathcal{P}_{R_i} \rho(X_{k_i}) + \epsilon_i
\end{align}
where $R_i \in SO(3)$ is the particle's orientation, $\mathcal{P}_{R_i}$ rotates and projects the density along the viewing direction, $t_i \in \mathbb{R}^2$ is an in-plane translation applied by $\mathcal{T}_{t_i}$, $\mathrm{PSF}_i$ is the point spread function, and $\epsilon_i$ is measurement noise. The Fourier transform of $\mathrm{PSF}_i$ is the contrast transfer function (CTF), $C_i$.

The image formation process causes ambiguity in structural inference. Differences between any two images may be due to molecular state, pose, imaging conditions, or measurement noise. Therefore, recovering biological information from cryo-EM images necessitates reasoning over different latent components of the same underlying generative process.

Many cryo-EM tasks can be viewed as different inverse problems over shared latent variables. Even when variables such as molecular structure, pose, and imaging conditions are not the target of the task, they confound an estimator. For example, estimating the pose $R_i$ requires determining which orientation of a candidate molecular structure is consistent with $y_i$. On the other hand, discriminating whether two particles arise from different molecular states requires distinguishing structural variation from differences in pose and imaging conditions. Reconstruction recovers $\rho(X_k)$ by combining particles according to both their orientations and their underlying structural states. Population estimation operates at the level of the structural ensemble, aggregating evidence about which molecular states are compatible with the observed particles in order to estimate $\pi(X_k)$.

This overlap in required information to solve a task motivates learning a common representation, $z_i$, with one encoder, $f_\theta(y_i)$ for every downstream task. Task-specific decoders and inference procedures can then re-organize learned concepts to solve each task. A useful representation should then preserve molecular information despite large variation in pose, imaging conditions, and noise, while remaining sensitive to biologically meaningful differences between molecular states. Moreover, these differences can be spatially localized. For example, a conformational change can be limited to a small region of a projected molecule. A global representation may discard small-scale variation in favor of larger elements of variation. 
We explore how to pose a learning task to create general protein representations such that they can be reused across diverse tasks and for any protein.
\section{Related Works}
\label{ref:related-works}

\paragraph{Learning solutions to cryo-EM inverse problems.}
Machine learning has been used to solve individual tasks in cryo-EM, from supervised particle picking with Topaz and crYOLO~\citep{bepler2019positive,wagner2019sphire} to direct structural inference.
CryoDRGN~\citep{zhong_cryodrgn_2021} represents density, conditioned on a latent conformation, with an implicit neural representation, while e2gmm~\citep{chen2021deep} represents heterogeneous structures with Gaussian mixtures; both assume known particle poses.
CryoDRGN2~\citep{zhong_cryodrgn2_2021} and CryoDRGN-AI~\citep{levy_cryodrgn-ai_2025} extend neural heterogeneous reconstruction to unknown poses, while CryoSPHERE~\citep{ducrocq_cryosphere_2025} models heterogeneity through approximately rigid motions of learned structural segments.
Other methods specialize in conformational inference, such as CryoSBI~\citep{dingeldein_amortized_2025}, or pose estimation and reconstruction with CryoFastAR~\citep{zhang_cryofastar_2025}.
These approaches demonstrate the value of learning for cryo-EM, but are generally built for a specific inference problem.
Our work aims to learn a representation that can be reused across proteins and downstream tasks.

\paragraph{Self-supervised representation learning.}
Modern self-supervised learning largely differs in how it defines agreement across views and what information the objective preserves.
Contrastive methods such as MoCo~\citep{he_momentum_2019} learn invariances by matching augmented views while contrasting different images.
DINO~\citep{caron_emerging_2021} removes explicit negatives through self-distillation and produces spatially meaningful patch features in vision transformers.
Masked-image methods such as MAE~\citep{he_masked_2021} reconstruct missing image content, while iBOT~\citep{zhou_ibot_2021} combines image-level self-distillation with masked prediction of patch representations.
DINOv2 and DINOv3~\citep{oquab_dinov2_2024,simeoni_dinov3_2025} scale this family of objectives to large datasets and models while retaining strong global and dense features.
We adopt this self-distillation paradigm, but construct the training views and invariances using the cryo-EM forward model.

\paragraph{Representation learning for cryo-EM.}
Self-supervised learning has been applied across the cryo-EM pipeline.
DRACO, CryoMAE, and cryo-EMMAE~\citep{shen_draco_2024,xu_cryomae_2024,zamanos_self-supervised_2025} learn from micrographs for particle localization and denoising, CryoLVM~\citep{fu_cryolvm_2026} learns representations of reconstructed density maps for restoration, and CryoSiam~\citep{stojanovska2025cryosiam} operates on tomographic data.
Closer to our setting, Cryo-IEF~\citep{yan_comprehensive_2026} pre-trains on picked single-particle images using a MoCo-style contrastive objective over augmented views, while CryoUNI~\citep{dai_cryo-em_2026} learns particle representations through paired denoising and reconstruction.
These methods show transfer beyond their pre-training objectives, but their evaluations largely remain within one or two stages of the pipeline or set of particle-level properties. However, Cryo-IEF evaluates structural classification, pose clustering, and image-quality assessment.
Our work further probes what biological information a general particle representation retains across conformational and compositional discrimination, population estimation, and particle grouping for reconstruction.

\section{Learning through the Cryo-EM Forward Model}
\label{sec:methods}

\begin{figure}[htbp!]
    \centering
    \includegraphics[width=\linewidth]{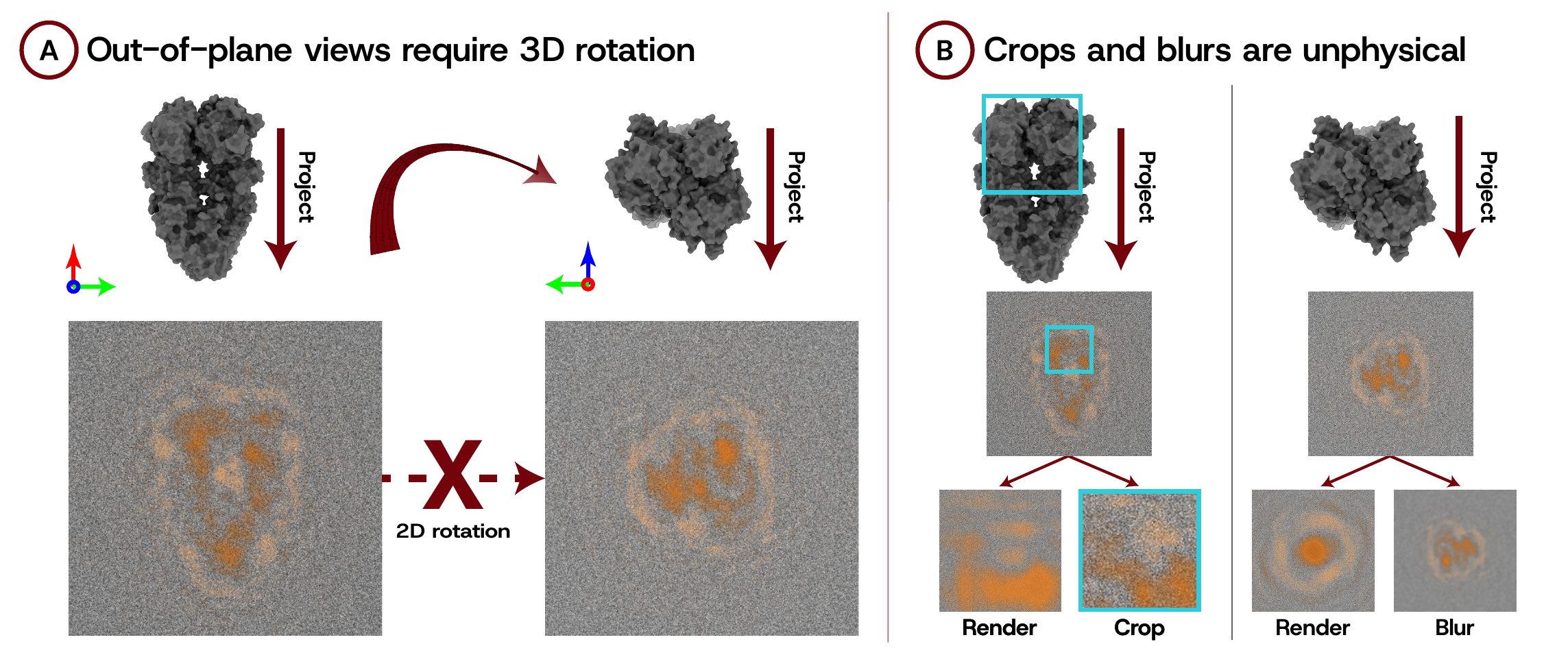}
    \caption{\textbf{Augmentations can be unphysical}. (A) 2D rotations of an image do not span all possible rotations of a 3D protein. (B) Image-space crops and blurs do not translate to the cryo-EM forward model.}
    \label{fig:augmentations-visualized}
\end{figure}

Many self-supervised learning methods use multiple views of the same observation as a training signal, using augmentations that alter image appearance while preserving semantic content.
When explicitly rendering new observations is impractical, as in natural scenes, image-space augmentations such as crops, rotations, color perturbations, and blur provide useful proxies for variation in the imaging process.
For example, a 2D rotation corresponds to an in-plane camera rotation, while Gaussian blur can approximate optical or motion blur.

The same strategy, however, is less natural for cryo-EM.
Transformations of an already projected particle image do not, in general, correspond to varying the latent variables of the cryo-EM forward model while holding structure fixed.
For example, rotating a particle image, as in Cryo-IEF, changes only its in-plane orientation and cannot generate a different out-of-plane projection (Figure~\ref{fig:augmentations-visualized}).
Similarly, Gaussian blur does not reproduce a change in defocus, whose effect is governed by the oscillatory contrast transfer function.
Thus, generic image-space augmentations may induce invariances that do not correspond to physically realizable variation in cryo-EM.

\looseness=-1
However, high-fidelity explicit forward models for cryo-EM already exist, such as CryoJAX~\citep{obrien_cryojax_2026}.
Instead of approximating changes to the image formation process with 2D augmentations, we can directly render alternate observations of the same structure:
\begin{equation}
    y_i \sim p(y \mid X_k, R_i, C_i, t_i, \ldots)
\end{equation}
while varying the latent quantities that determine how a fixed molecular structure is observed.
We engineered CryoJAX to render these views fast enough to run inside the training loop.
This lets us train models with SSL with grounded positive pairs without the cost generally associated with rendering.

Given this, the remaining question is which SSL objective should learn from these pairs.
Contrastive objectives additionally require defining negative examples, which is difficult in cryo-EM.
Two different particle images can arise from the same molecular structure, while their visual similarity or dissimilarity may instead be a function of their pose or imaging conditions.
We use DINOv3~~\citep{simeoni_dinov3_2025} for image-level self-distillation and complement it with the iBOT masked-patch objective~\citep{zhou_ibot_2021} to supervise spatially resolved features without needing explicit negative pairs.

\section{Pre-training Data and View Construction}
\label{sec:data-view-construction}

Constructing the targets described in the previous section at scale presents a practical tradeoff.
Simulated images provide precise control over structure and imaging conditions, but cannot capture the full acquisition statistics of experimental data; experimental particles provide this realism, but their underlying structures and correspondences are unknown.
We pre-train on simulated data with marginal amounts of experimental images.
Afterwards we introduce equal amounts of experimental and simulated data during a mid-training phase, using a different view-construction strategy for each.

\paragraph{Simulated and experimental training data.}
For simulated data, we use OpenProteinSet, a processed snapshot of the PDB~\citep{berman_protein_2000, ahdritz_openproteinset_2023} released as part of OpenFold~\citep{ahdritz_openfold_2024}.
After filtering, our training set consists of 268k molecular assemblies determined across a range of experimental modalities, including X-ray crystallography and cryo-EM.
We render particle images online using CryoJAX~\citep{obrien_cryojax_2026}, independently sampling poses, translations, CTF parameters, noise levels, and \AA/px values during training.
We curate 51M picked single-particle images from 214 EMPIAR~\citep{iudin_empiar_2023} entries for the mid-training phase.
The full filtering and preprocessing procedure for both data sources is described in Appendix~\ref{appendix:dataset-processing}.

\paragraph{View Construction.}
The two data sources provide different information about correspondence between particle images.
For simulated data, the underlying structure $X_k$ is known, so we construct paired views by rendering the same $X_k$ under independently sampled imaging conditions and noise, and with probability $p=0.5$ independently sample the pose of each view.
With this, we are able to retain exact structural correspondence even when two views represent substantially different projections of the molecule.

For experimental particles, the underlying $X_k$ is unknown and exact structural pairs cannot be constructed.
Instead, with probability $p=0.5$ we pair a particle with another particle from the same EMPIAR entry.
This is a weaker correspondence than we can achieve with synthetic data.
Particles from the same entry generally arise from the same molecular system, but may differ in conformation or composition.
Overall, these datasets are complementary.
Simulated views provide strong structural correspondence, while experimental views expose the model to realistic variation that can be absent from the simulator.

\paragraph{Mixing simulated and experimental data.}
We find that introducing experimental data at the start of training destabilizes optimization and can lead to representation collapse.
Instead, we first pre-train on simulated data with marginal amounts of experimental images and gradually introduce experimental particles.
The first 50k steps of training use a 93.75:6.25 ratio between simulated particles and experimental particles.
Over the next 50k steps of training the experimental fraction is increased on a cosine schedule to a stochastic 50:50 mixture, which is then used for the remainder of training.

\paragraph{Pre-training objective and architecture.}
We instantiate the learning procedure described in the previous section using DINOv3 with the iBOT masked-patch objective.
The student and EMA teacher are vision transformers with a patch size of $16\times16$ pixels, a learnable \verb|[CLS]| token, and four register tokens~\citep{darcet_vision_2024}.
We train ViT-S (21.4M), ViT-B (85.3M), ViT-L (304.2M), and ViT-g (1.1B) variants, collectively called the EMber family.
Complete loss definitions, architectural details, regularization terms, and optimization hyperparameters are provided in Appendix~\ref{appendix:training-details}.
\section{Evaluating What~\model~Learns}
\label{sec:evaluations}

We aim to evaluate~\model~comprehensively so we can understand what information is accessible from the learned representations, how that information can be reused, and ultimately, performance on downstream tasks.
Our evaluations span non-parametric use of frozen features, linear probes, full fine-tuning, and integration into a reconstruction pipeline.
First, we use experimental data to assess if frozen representations contain molecular information and if that information is global or spatially distributed.
Then, we test how the representation can be adapted to discriminate between subtle conformational changes, if representations learned at an image level can be aggregated to recover conformational populations, and whether the learned representation can be reused for particle grouping during reconstruction.
For each evaluation, additional details and hyperparameters are provided in Appendix~\ref{appendix:evaluations}.

\subsection{Probing Structural Information in Experimental Images}
\label{sec:cryoppp-knn}

We first test if the representations learned by \model~can organize experimental particles consistently enough to recover dataset identity without learning task-specific readouts.
Following prior work, we build a $k$-nearest neighbor classifier over 29 experimental particle datasets from CryoPPP~\citep{dhakal_cryoppp_2023}.
As shown in Table~\ref{tab:cryoppp-knn}, models are able to easily discriminate between EMPIAR datasets in latent space.~\model~ViT-S reaches 92.17\% Top-1 accuracy and 99.77\% Top-5 accuracy, with larger models saturating quickly.
At comparable model sizes,~\model~improves over Cryo-IEF and CryoUNI.
Particle images from apo (EMPIAR-10005) and ligand-bound (EMPIAR-10059) TRPV1 datasets occupy overlapping regions in dimensionality reduction plots (Appendix~\Fref{fig:cryoppp-trpv1}), consistent with their shared molecular identity.

% \begin{table}[htbp!]
%     \centering
%       \begin{tabular}{lrr}
%         \toprule
%         Model & Top-1 (\%) & Top-5 (\%) \\
%         \midrule
%         \rowcolor{paper-deep} \model-S & 92.17 & 99.77 \\
%         \rowcolor{paper-deep} \model-B & 94.47 & \textbf{99.94} \\
%         \rowcolor{paper-deep} \model-L & 93.63 & 99.60 \\
%         \rowcolor{paper-deep} \model-G & \textbf{95.77} & 99.86 \\
%         \rowcolor{paper} Cryo-IEF-S & 81.78 & 98.59 \\
%         \rowcolor{paper} Cryo-IEF-B & 81.12 & 98.33 \\
%         \rowcolor{paper} Cryo-UNI-S & 82.68 & 98.79 \\
%         \rowcolor{paper} Cryo-UNI-B & 85.88 & 98.98 \\
%         \bottomrule
%       \end{tabular}
%     \caption{\textbf{$k$-nearest neighbor EMPIAR dataset discrimination.} Top-1 and Top-5 accuracies (\%) across 29 experimental
%             datasets from CryoPPP. High classification accuracy may reflect both molecular structure and acquisition-specific information.}
%     \label{tab:cryoppp-knn}
% \end{table}

However, this result is difficult to interpret biologically.
EMPIAR datasets vary the molecule being imaged and acquisition-specific factors.
This experiment shows that the representation captures strong and reproducible signal in experimental particles, but not which part of that signal is structural.
Next, we design a more controlled experimental setting to isolate molecular composition from imaging conditions.

\paragraph{Compositional discrimination under controlled imaging conditions.}
\label{sec:cahra-probe}

\begin{figure}[htbp!]
    \centering
    \includegraphics[width=1.0\linewidth]{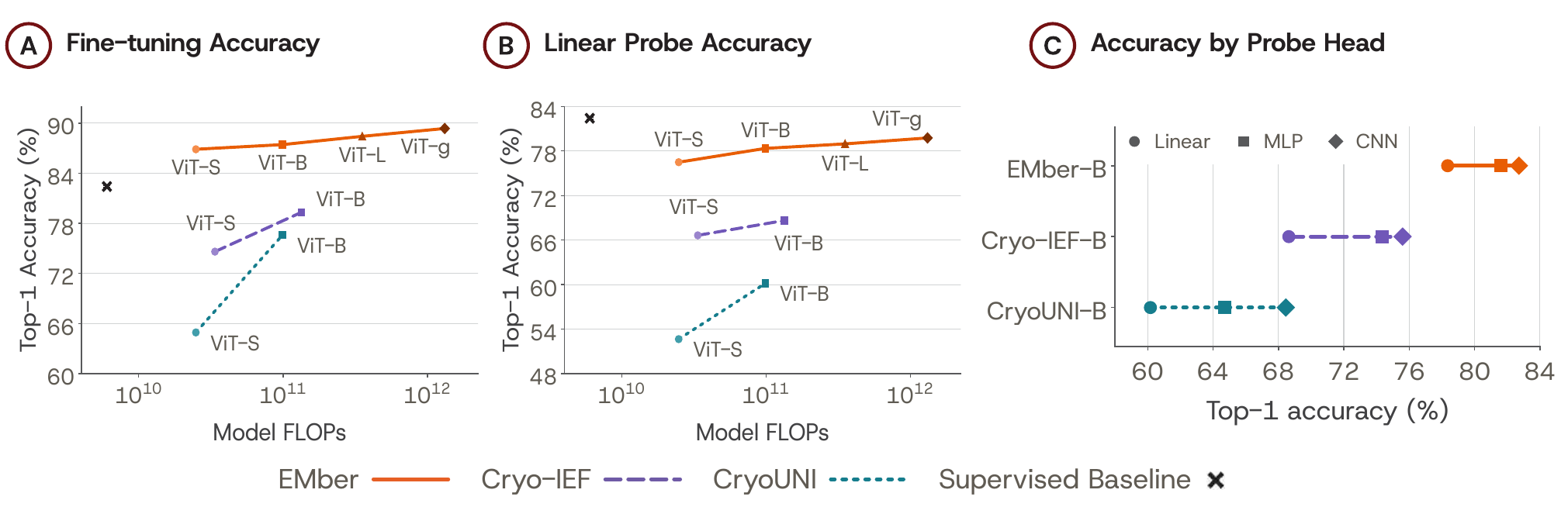}
    \caption{\textbf{Compositional heterogeneity classification on CAHRA experimental data.}
    (A) Fine-tuning accuracy versus model FLOPs, with a supervised ResNet-18 reference.
    (B) Accuracy versus model FLOPs when linear probing frozen backbones, with a reference supervised ResNet-18.
    (C) Pooled linear/MLP readouts with a convolutional head that preserves spatial information.}
    \label{fig:cahra-probe-comparison}
\end{figure}

To directly test whether frozen representations retain molecular information, we use the CAHRA compositional heterogeneity challenge~\citep{feathers_cahra_2026} which contains labeled particles from six proteins collected under similar microscopy conditions which are not seen during pre-training.
We fine-tune the pre-trained backbones for this task, as well as multiple probes on frozen features.
On pooled features, we train linear and MLP probes, and on the unpooled two-dimensional patch features we train a convolutional probe. Full details are in Appendix~\ref{appendix:evaluations}.

The results when fine-tuning the encoders is shown in~\Fref{fig:cahra-probe-comparison}A.
Across all model sizes~\model~is the only model which outperforms the fully-supervised baseline ResNet-18. 
Fig.~\ref{fig:cahra-probe-comparison}B shows that compositional information is already linearly accessible from frozen representations.~\model-g reaches 79.76\% with a linear probe, only 2.65 percentage points below the fully supervised ResNet-18, and outperforms Cryo-IEF and CryoUNI at comparable model sizes.
Fig.~\ref{fig:cahra-probe-comparison}C tests whether pooling discards useful information.
Across all pre-trained backbones, the parameter-matched convolutional probe outperforms the MLP.~\model-B reaches 82.72\%, comparable to the supervised baseline.
As the trend is similar for Cryo-IEF and CryoUNI, this result supports retaining the spatial organization of patch features rather than attributing the gain solely to our patch-level objective.

%%%% Raw numbers for reference
% Left panel — linear-probe accuracy vs. estimated backbone GFLOPs/image:
% - Ours ViT-S: 24.864 GFLOPs; 76.4933 ± 0.0157%
% - Ours ViT-B: 99.075 GFLOPs; 78.3539 ± 0.0609%
% - Ours ViT-L: 353.440 GFLOPs; 78.9628 ± 0.0336%
% - Ours ViT-G: 1317.349 GFLOPs; 79.7628 ± 0.0652%
% - Cryo-IEF ViT-S: 11.035 GFLOPs; 73.3661 ± 0.0091%
% - Cryo-IEF ViT-B: 43.898 GFLOPs; 75.3578 ± 0.0291%
% - Cryo-UNI ViT-S: 2.952 GFLOPs; 61.5133 ± 0.0301%
% - Cryo-UNI ViT-B: 11.764 GFLOPs; 71.6478 ± 0.0272%
% - Supervised baseline (ResNet-18): 6.017 GFLOPs; 82.4167 ± 0.1341%

% Right panel — accuracy by probe head:
% - Ours ViT-B:
%   - Linear: 78.3539 ± 0.0609%
%   - MLP: 81.6128 ± 0.0940%
%   - CNN: 82.7189 ± 0.1994%
% - Ours ViT-G:
%   - Linear: 79.7628 ± 0.0652%
%   - MLP: 81.4367 ± 0.0177%
%   - CNN: 82.7378 ± 0.0772%
% - Cryo-IEF ViT-B:
%   - Linear: 75.3578 ± 0.0291%
%   - MLP: 79.0572 ± 0.1048%
%   - CNN: 80.4883 ± 0.0481%
% - Cryo-UNI ViT-B:
%   - Linear: 71.6478 ± 0.0272%
%   - MLP: 72.0094 ± 0.1050%
%   - CNN: 72.6728 ± 0.0256%

\subsection{Adapting Representations to Conformational Variation}
\label{sec:hsp-finetune}

\begin{figure}[htbp]
    \centering
    \includegraphics[width=\linewidth]{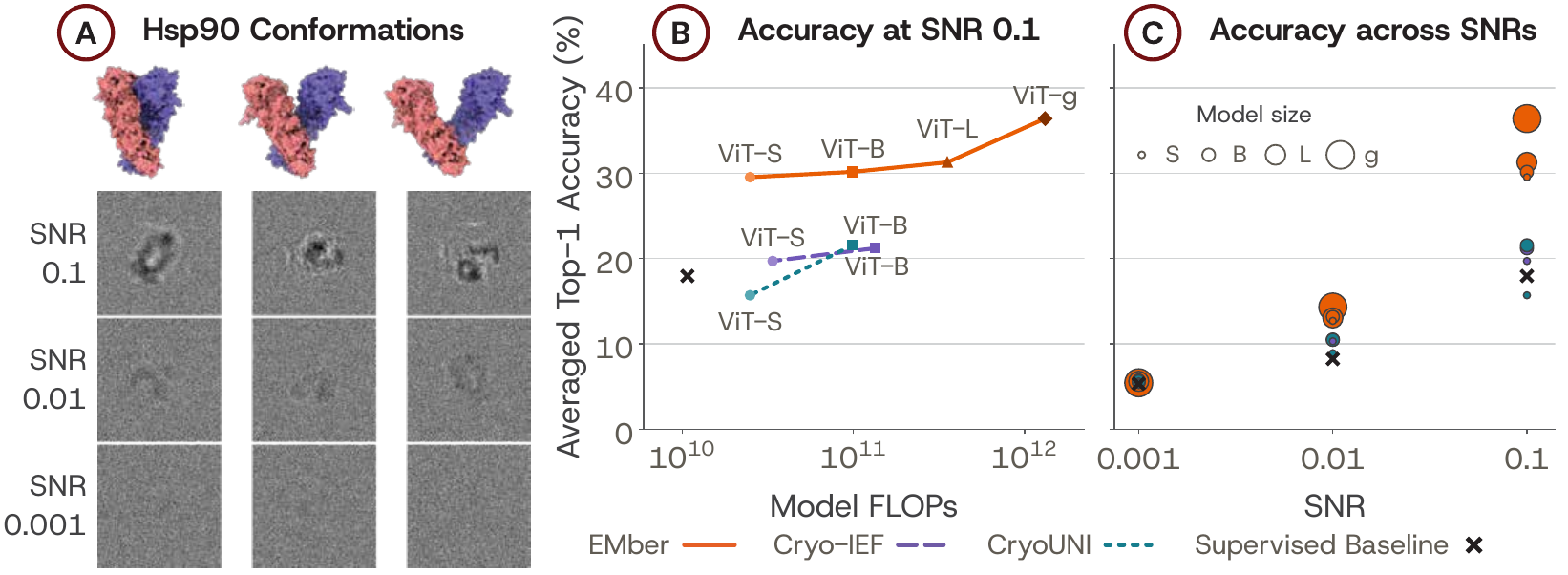} 
    \caption{\textbf{Fine-tuning for Hsp90 conformation classification.}
    We discretize Hsp90 conformations according to CMA and CMB chain-opening angles, each with 20 classes, and average accuracy across the two prediction heads (random chance: 5\%).
    (A) Representative conformations and rendered particles at different SNRs.
    (B) Top-1 accuracy at SNR 0.1 versus model forward FLOPs.
    (C) Accuracy across SNRs. Model differences largely disappear at SNR 0.001, where all methods approach chance.
    }
    \label{fig:hsp90-scaling}
\end{figure}

The previous experiments ask what molecular information can be extracted from frozen representations.
We now evaluate whether pre-training provides a useful initialization when the full representation is adapted to distinguish subtle conformational changes within the same molecule.
Since pose is unknown, this is a difficult task. 
Methods based on template search or simulation-based inference require dense coverage of pose space for training~\citep{giammar_leopard-em_2025,dingeldein_amortized_2025}.
We purposefully evaluate in an under-sampled pose regime.

We study heat shock protein 90 (Hsp90), which is a standard system for studying conformational heterogeneity~\citep{seitz_simulation_2019,dingeldein_amortized_2025}.
Two opening angles, CMA and CMB, parameterize the transition from open to closed states.
We discretize this continuous transition spectrum into 20 bins, yielding 400 conformations, and jointly fine-tune the backbone with one readout head per angle.
The training set contains 200,000 images---only 500 per conformation---so each state is observed under a sparse set of poses.
We create validation sets at SNR 0.1, 0.01, and 0.001 which measure how performance varies as signal is removed.

%%% raw numbers for reference
% middle panel — mean top-1 accuracy at SNR 0.1:
% - Ours ViT-S: 24.864 GFLOPs; 29.5190%
% - Ours ViT-B: 99.075 GFLOPs; 30.1430%
% - Ours ViT-L: 353.440 GFLOPs; 31.2600%
% - Ours ViT-g: 1317.349 GFLOPs; 36.3880%
% - Cryo-IEF ViT-S: 11.035 GFLOPs; 20.3780%
% - Cryo-IEF ViT-B: 43.898 GFLOPs; 22.0740%
% - Cryo-UNI ViT-S: 2.952 GFLOPs; 10.1000%
% - Cryo-UNI ViT-B: 11.764 GFLOPs; 18.5570%
% - Supervised ResNet-18 baseline (horizontal line): 17.9450%

At SNR 0.1, nearly every pre-trained model outperforms the ResNet-18 trained from scratch, demonstrating that pre-training excels when task-specific data sparsely sample pose space (\Fref{fig:hsp90-scaling}B).
\model-S reaches 29.52\% accuracy compared with 17.95\% for the ResNet-18 baseline, while scaling to~\model-g raises accuracy to 36.39\%.
The full model comparison is shown in \Fref{fig:hsp90-scaling}B.

However, representation quality is limited by the epistemic identifiability of noisy images.
As SNR decreases, all models degrade sharply and approach chance at SNR 0.001 (\Fref{fig:hsp90-scaling}C).
Thus, pre-training makes fine-grained conformational information easier to learn from sparse pose coverage, but cannot recover structural signal once the observations themselves are noise-dominated.
% \pc{I recommend putting in SI results when training with more images per conformation. We should mention that we see the same trends for model size and SNR but that accuarcy improves overall - which is what is expected when training with more images.}
% \nc{Made a note to run in SI, and on my radar}
% unlikely I will get to doing the 1M eval :'( we will add after the deadline

\subsection{From Particle Representations to Conformational Populations}
\label{sec:spike-pop-est}

% ORIGINAL TABLE AND FIG
% \begin{table}[htbp!]
%     \centering
%       \begin{tabular}{lrr}
%         \toprule
%         Model & Top-1 (\%) & Top-5 (\%) \\
%         \midrule
%         \rowcolor{paper-deep} \model-S & 92.17 & 99.77 \\
%         \rowcolor{paper-deep} \model-B & 94.47 & \textbf{99.94} \\
%         \rowcolor{paper-deep} \model-L & 93.63 & 99.60 \\
%         \rowcolor{paper-deep} \model-G & \textbf{95.77} & 99.86 \\
%         \rowcolor{paper} Cryo-IEF-S & 81.78 & 98.59 \\
%         \rowcolor{paper} Cryo-IEF-B & 81.12 & 98.33 \\
%         \rowcolor{paper} Cryo-UNI-S & 82.68 & 98.79 \\
%         \rowcolor{paper} Cryo-UNI-B & 85.88 & 98.98 \\
%         \bottomrule
%       \end{tabular}
%     \caption{\textbf{$k$-nearest neighbor EMPIAR dataset discrimination.} Top-1 and Top-5 accuracies (\%) across 29 experimental
%             datasets from CryoPPP. High classification accuracy may reflect both molecular structure and acquisition-specific information.}
%     \label{tab:cryoppp-knn}
% \end{table}
% \begin{figure}
%     \centering
%     \includegraphics[width=0.5\linewidth]{figures/spike-pop-est.pdf}
%     \caption{The plot shows estimated minus true Spike-Up population in the mixed-SNR setting. Each point is an average across three probe training seeds and 10 population sampling seeds, with shaded regions showing $\pm1$ standard deviation.}
%     \label{fig:synth-spike-pop-est}
% \end{figure}

\begin{figure}[htbp!]
    \centering
    % Figure on the left
    \begin{minipage}[t]{0.5\textwidth}
        \vspace{0pt}
        \centering
        \includegraphics[width=\linewidth]{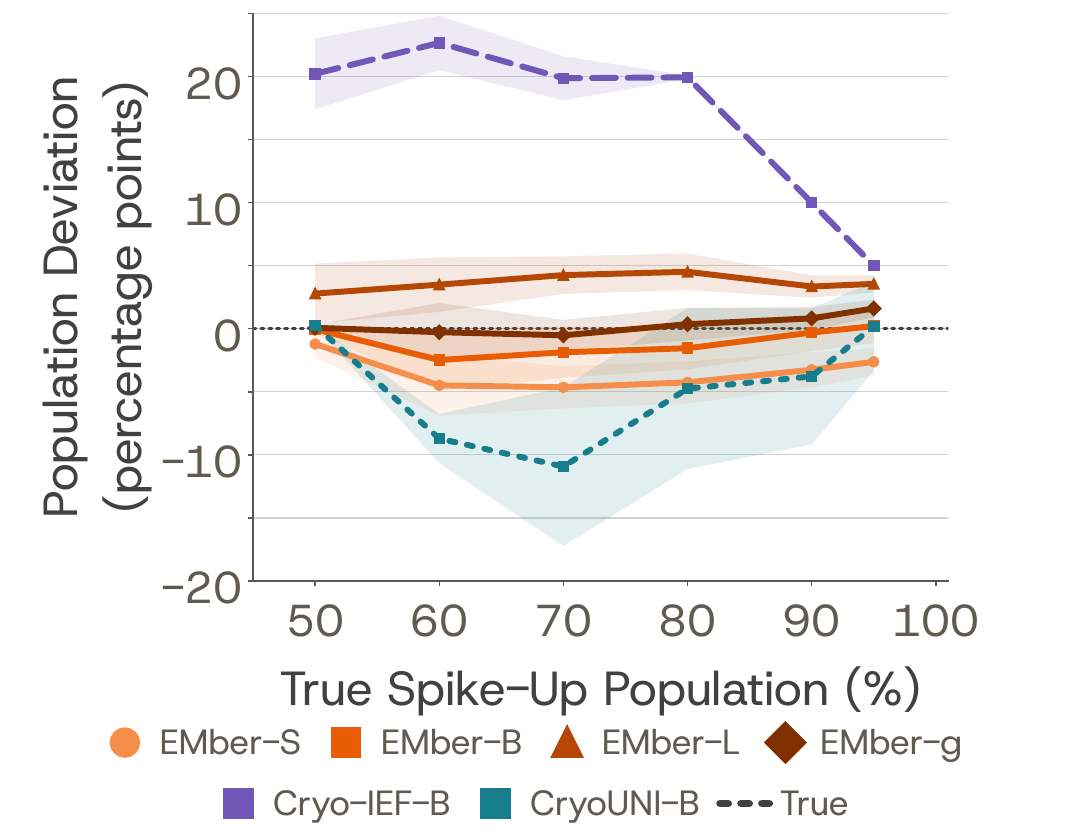}
        \caption{The plot shows estimated minus true Spike-Up population in the mixed-SNR setting. Each point is an average across three probe training seeds and 10 population sampling seeds, with shaded regions showing $\pm1$ standard deviation.}
        \label{fig:synth-spike-pop-est}
    \end{minipage}
    \hfill
    % Table on the right
    \begin{minipage}[t]{0.48\textwidth}
        \vspace{0pt}
        \centering
        \begin{tabular}{lrr}
            \toprule
            Model & Top-1 (\%) & Top-5 (\%) \\
            \midrule
            \rowcolor{paper-deep} \model-S & 92.17 & 99.77 \\
            \rowcolor{paper-deep} \model-B & 94.47 & \textbf{99.94} \\
            \rowcolor{paper-deep} \model-L & 93.63 & 99.60 \\
            \rowcolor{paper-deep} \model-g & \textbf{95.77} & 99.86 \\
            \rowcolor{paper} Cryo-IEF-S & 81.78 & 98.59 \\
            \rowcolor{paper} Cryo-IEF-B & 81.12 & 98.33 \\
            \rowcolor{paper} CryoUNI-S & 82.68 & 98.79 \\
            \rowcolor{paper} CryoUNI-B & 85.88 & 98.98 \\
            \bottomrule
        \end{tabular}
        \captionof{table}{\textbf{$k$-nearest neighbor EMPIAR dataset discrimination.} Top-1 and Top-5 accuracies (\%) across 29 experimental datasets from CryoPPP. EMber-S outperforms the ViT-B variants of both baselines on Top-1 accuracy. Further, Top-1 accuracy generally gets better as a function of model size across both~\model, Cryo-IEF, and Cryo-UNI. ~\model~performs approximately 0.86pp better than Small and Base over baselines, on average. High classification accuracy may reflect both molecular structure and acquisition-specific information.}
        \label{tab:cryoppp-knn}
    \end{minipage}
\end{figure}

Particle-level classification is not always the quantity of biological interest.
Since both pose and noise can make individual images ambiguous~\citep{evans_counting_2025}, we measure whether representations across many particles can be aggregated to recover a property of the structural ensemble.
Specifically, we estimate the conformational population distribution $\pi(X_k)$ from image-level predictions.

We use the SARS-CoV-2 Spike molecular dynamics dataset from CryoBench~\citep{wieczor_omicron_2023,jeon_cryobench_2024} and group the conformational ensemble into Spike-Up and Spike-Down states (Appendix~\Fref{fig:spike-cv-filtering}).
We train a linear classifier on frozen features using 50k simulated images, calibrate on held-out data, and pass the probe's outputs to the expectation-maximization procedure from~\citep{evans_counting_2025}. Since a linear probe's logits are not calibrated likelihoods, we perform Platt scaling on a held-out calibration set before the EM procedure.
The evaluation populations contain images spanning SNR 0.1, 0.01, and 0.001. Results stratified by SNR, additional training sizes, and the complete EM setup are provided in Appendix~\ref{appendix:spike-pop-est}.

%%% Numbers for reference for writing
% Population MAE — mean ± SD, in percentage points:
% - Ours ViT-S: 3.4136 ± 0.4859
% - Ours ViT-B: 1.5915 ± 0.2598
% - Ours ViT-L: 3.6652 ± 0.6330
% - Ours ViT-g: 1.1069 ± 0.2504
% - Cryo-IEF ViT-B: 16.2753 ± 0.6085
% - CryoUNI ViT-B: 5.6393 ± 1.7312

Population estimates from evaluated models reveal a larger separation between representations than particle-level accuracy alone would suggest.
In our mixed-SNR setting,~\model-B reaches a mean absolute population error of $1.59 \pm 0.26$ percentage points, compared with $16.28 \pm 0.61$ for Cryo-IEF-B and $5.64 \pm 1.73$ for CryoUNI-B (\Fref{fig:synth-spike-pop-est}).
~\model-g reaches 1.11 percentage points, although this scaling trend is not consistent across the other training set sizes reported in Appendix~\ref{appendix:spike-pop-est}.
Cryo-IEF systematically overestimates the Spike-Up fraction while CryoUNI underestimates it. Across all tested mixtures,~\model~remains much closer to the true population.
These results demonstrate that representations derived from our pre-training setup can be useful even when individual particles are ambiguous, and individually weak image-level evidence can be aggregated to recover an ensemble-level quantity.

\subsection{Representations for Heterogeneous Reconstruction}
\label{sec:synthetic-reconstruction}

Finally, we measure whether representations learned with our objective can be reused inside a coupled cryo-EM inverse problem.
Heterogeneous reconstruction requires estimating both particle pose and grouping particles by structure. Since reconstruction quality by itself does not reveal which part of the inference failed, we design a two-phase protocol to isolate particle grouping.

In Phase 1, we run CryoDRGN-Ab-initio~\citep{levy_cryodrgn-ai_2025} with unknown poses while replacing its encoder with~\model~or a baseline encoder.
In Phase 2, we cluster the learned latent space with k-means++~\citep{arthur_k-means_2007,pedregosa_scikit-learn_2011} and reconstruct each group in RELION~\citep{scheres_relion_2012} with ground truth poses.
As pose is unknown during pre-training and fixed during the final reconstruction, differences in reconstruction quality reflect how well each representation organizes particles by structure.

% \begin{wrapfigure}{R}{0.5\textwidth}
\begin{figure}[htbp!]
    \centering
    \includegraphics[width=\linewidth]{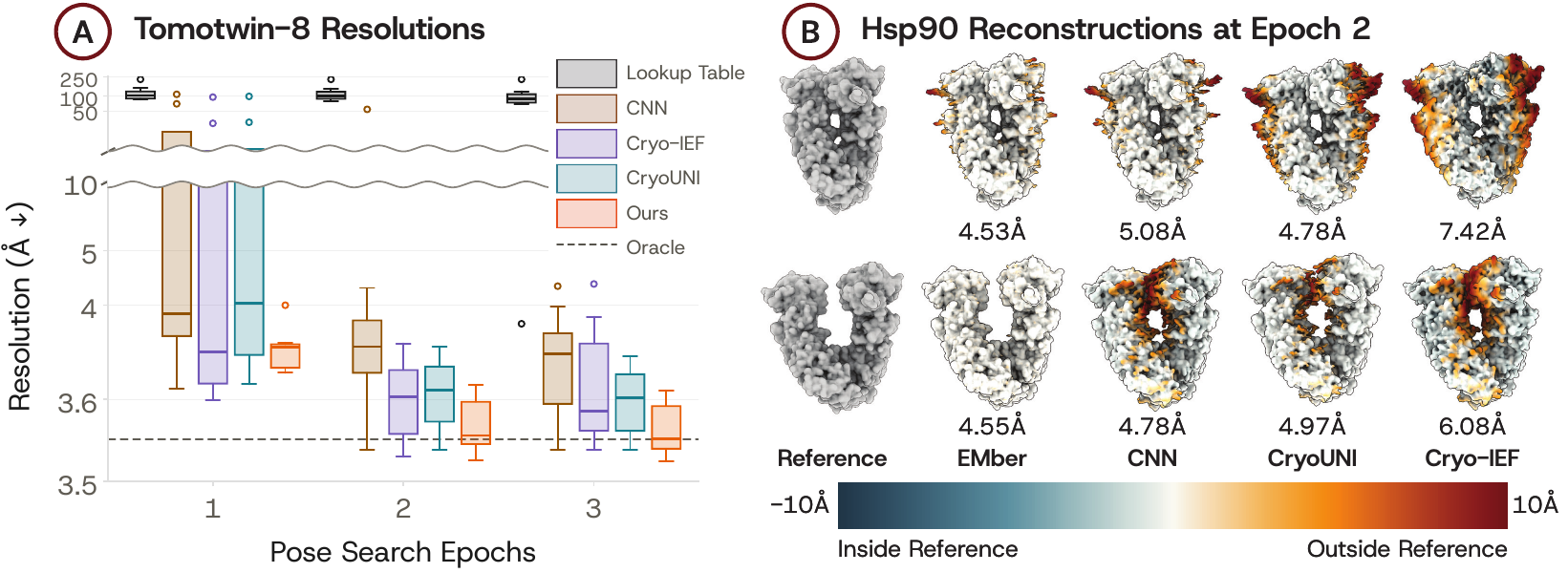}
    \caption{\textbf{Particle grouping measured through reconstruction resolution.}
    After each epoch of CryoDRGN-AI pose search, particles are clustered in the learned latent space and reconstructed using ground truth poses. 
    (A) Tomotwin-8: Boxes show the spread in FSC=0.5 resolution across eight structures.
    The dashed line is an Oracle reconstruction obtained with ground truth group assignments. Notably,~\model~is already near the Oracle after one epoch.
    \model~reaches near reference resolution within two epochs, and has less variation across the reconstructed clusters compared to baseline methods.
    (B) Hsp90: Reconstructions visualized after 2 epochs of CryoDRGN-AI pose search.
    Colors represent the signed distance to the reference surface (negative inside, positive outside).
    }
    \label{fig:tomotwin-synth-recon-res}
\end{figure}
% \end{wrapfigure}

% \paragraph{Setup and Baselines.} 
We compare frozen ViT-B backbones from~\model,Cryo-IEF, and CryoUNI with the default CryoDRGN CNN encoder and its lookup table baseline.
All methods use the same CryoDRGN-Ab-initio pose-search procedure, latent dimensionality, particle clustering, and RELION reconstruction protocol.
Oracle reconstructions use ground truth structure assignments and poses.
Complete details on head architectures, clustering settings, and FSC curves are provided in Appendix~\ref{appendix:synthetic-reconstruction}.

%%% Resolution numbers for writing reference:
% TomoTwin — 8 reconstructed clusters per method/epoch:
% - Lookup Table 
%   - Epoch 1: 222.9345 ± 0.0389
%   - Epoch 2: 182.9609 ± 55.0650
%   - Epoch 3: 130.9213 ± 82.5956
% - CNN 
%   - Epoch 1: 29.9322 ± 52.5553
%   - Epoch 2: 10.0384 ± 17.7923
%   - Epoch 3: 3.7589 ± 0.2357
% - Cryo-IEF 
%   - Epoch 1: 21.0358 ± 39.8662
%   - Epoch 2: 3.6203 ± 0.0838
%   - Epoch 3: 3.7105 ± 0.2527
% - Ours 
%   - Epoch 1: 3.7400 ± 0.1128
%   - Epoch 2: 3.5689 ± 0.0342
%   - Epoch 3: 3.5639 ± 0.0314
%
% - CryoUNI: cls_mlp
%   - Epoch 1: 15.4264 ± 21.6128
%   - Epoch 2: 3.7746 ± 0.1584
%   - Epoch 3: 3.7353 ± 0.1261
% - CryoUNI: spatial_dense
%   - Epoch 1: 21.2370 ± 30.2175
%   - Epoch 2: 3.6267 ± 0.0720
%   - Epoch 3: 3.6110 ± 0.0609

% HSP90 — 2 reconstructed clusters per method/epoch:
% - Lookup Table
%   - Epoch 1: 8.9125 ± 0.2896
%   - Epoch 2: 8.9393 ± 0.2258
%   - Epoch 3: 8.7808 ± 0.2634
% - CNN 
%   - Epoch 1: 6.8501 ± 0.5917
%   - Epoch 2: 4.9292 ± 0.2173
%   - Epoch 3: 4.7828 ± 0.0508
% - Ours 
%   - Epoch 1: 7.0656 ± 0.6988
%   - Epoch 2: 4.5409 ± 0.0139
%   - Epoch 3: 4.5227 ± 0.0133
% - CryoUNI: cls_mlp
%   - Epoch 1: 5.7477 ± 0.2184
%   - Epoch 2: 5.2480 ± 0.2625
%   - Epoch 3: 5.2446 ± 0.2338
% - CryoUNI: spatial_dense
%   - Epoch 1: 5.3780 ± 0.0128
%   - Epoch 2: 4.8718 ± 0.1342
%   - Epoch 3: 4.7614 ± 0.0191

%          \rowcolor{paper} Cryo-IEF (ViT-B)
          % & \textbf{6.43} & \textbf{5.46}
          % & 7.42 & 6.08 \\

\paragraph{Tomotwin-8.}
We evaluate compositional heterogeneity using eight structures from the CryoBench Tomotwin system~\citep{rice_tomotwin_2023,jeon_cryobench_2024}.
After a single pose-search epoch,~\model~already produces clusters that reconstruct to $3.74\pm0.11$~\AA---only 0.18~\AA~from the Oracle resolution of 3.56~\AA~(\Fref{fig:tomotwin-synth-recon-res}).
By epoch two, it reaches $3.57\pm0.03$~\AA, effectively matching the reference.
Cryo-IEF reaches comparable resolution after two epochs, while the CNN encoder requires three. The full distribution across reconstructed structures can be seen in \Fref{fig:tomotwin-synth-recon-res}A.
These results shows that the~\model~pre-training setup produces a representation that is already close to the underlying compositional grouping without any task-specific supervision over structural variables.

% \begin{figure}[htbp!]
%     \centering
%     \includegraphics[width=\linewidth]{figures/hsp90-signed-distance-layout-a.pdf}
%     \caption{\textbf{Reconstructed Hsp90 open and closed states.}
%     Maps are reconstructed from particle groups obtained after CryoDRGN-Ab-initio pose search.
%     By epoch 2,~\model~separates the two states well enough to recover the open gap between the Hsp90 arms at near-Oracle resolution.
%     \nc{block of sections; drop lookup table}}
%     \label{fig:hsp-synth-reconstruction}
% \end{figure}

\paragraph{Hsp90.}
Finally, we revisit the open and closed Hsp90 states from Section~\ref{sec:hsp-finetune}.
This provides a conformational counterpart to Tomotwin-8, where the structural difference is more localized.
After two epochs,~\model~reconstructs the closed and open states at 4.53~\AA~and 4.55~\AA, respectively, within 0.05~\AA~of the corresponding Oracle resolutions.
Although CryoUNI produces the best reconstruction after the first epoch, its grouping does not converge to the Oracle solution by epoch two, visualized in~\Fref{fig:tomotwin-synth-recon-res}B.
All methods recover the closed state while only~\model~recovers the open state at near-reference resolution while visibly resolving the gap between the two Hsp90 arms.

Across both compositional and conformational heterogeneity, representations learned with~\model~organize particles in a way that can be directly reused by a downstream reconstruction procedure with minimal re-training.
\section{Conclusion}
\label{sec:conclusion}

We introduce~\model, a family of general cryo-EM representations learned with self-supervision through the cryo-EM forward model.
Our evaluations provide a framework for studying what cryo-EM representations retain: distinguishing dataset discrimination from evidence of molecular information, comparing global and spatial readouts, and testing whether those representations support downstream structural inference.
Across frozen probing, fine-tuning, population estimation, and reconstruction,~\model~sets a new performance-compute frontier across several structural inference problems, while our analyses show that biologically relevant information is retained in both global and spatial features.
Our results provide a practical recipe for scaling representation learning on cryo-EM data: use the forward model to construct physically meaningful supervision, combine structurally precise simulated data with diverse experimental particles, and reuse the resulting representation across downstream analyses.
There remains substantial room to improve how simulated and experimental data are mixed, how spatial features are supervised under extreme noise, and how much experimental diversity can be recovered from currently deposited particle data.
Nevertheless, the breadth of transfer we observe suggests that representation learning on noisy cryo-images alone can learn to solve multiple inverse problems without explicit supervision, and opens the door to extending the same approach to micrographs, tomography, and other structural inference problems.
\section{Acknowledgments}

The authors were supported in part by the U.S. Department of Energy, Office of Science, Office of Advanced Scientific Computing Research, via the RAPIDS Institute under the Scientific Discovery through Advanced Computing (SciDAC) program; the U.S. Department of Defense; The House Fund; the Simons Foundation; and the State of Maryland.
The Flatiron Institute is a division of the Simons Foundation. 
The authors would like to thank the Scientific Computing Core, a division of the Flatiron Institute, and the U.S. Department of Defense's High Performance Computing Modernization Program for computational support.
We also thank Luke Evans, Michael O'Brien, David Silva-S\'anchez, Kithmini Herath, Henry Smith, and members of the ASK Group for their helpful feedback and discussions.

% \clearpage

\bibliography{main}
\bibliographystyle{plainnat}
% \input{post-script}

%%%%%%%%%%%%%%%%%%%%%%%%%%%%%%%%%%%%%%%%%%%%%%%%%%%%%%%%%%%%
\clearpage
\appendix
\section*{Appendices}
\section{Dataset Processing}
\label{appendix:dataset-processing}

\begin{figure}[htbp!]
    \centering
    \includegraphics[width=1\linewidth]{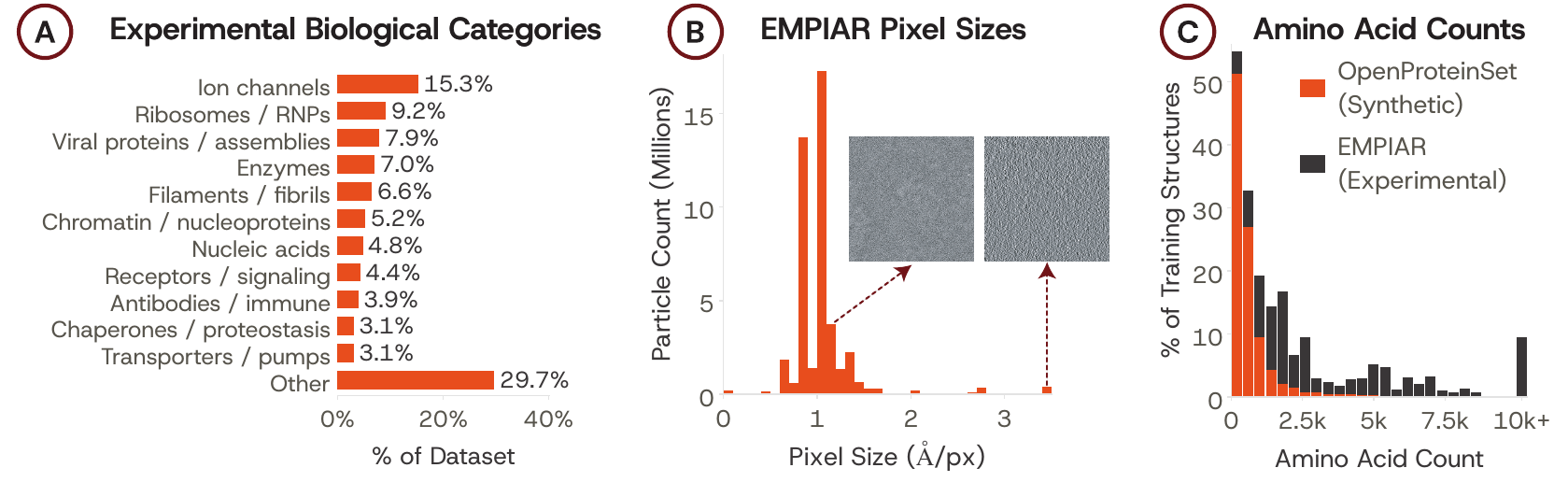}
    \caption{\textbf{Dataset Composition.} 
    Visualization of statistics coming from our EMPIAR dataset and the OpenProteinSet containing assemblies. 
    (A) Training set EMPIAR entries organized by coarse biological categories.
    (B) Pixel sizes of EMPIAR entries. At the same pixel dimensions, larger pixel sizes correspond to wider field of views.
    (C) Amino acid counts of synthetic and experimental datasets.
    }
    \label{fig:dataset-composition}
\end{figure}

\paragraph{EMPIAR Processing.} We take a snapshot of deposited entries within EMPIAR as of \verb|June 10, 2026| and apply a series of processing steps. 
We download particles from the following EMPIAR categories: 
\begin{enumerate}
    \item Picked particles, single frame, processed
    \item Picked particles, multiframe, processed
    \item Picked particles, single frame, unprocessed
    \item Picked particles, multiframe, unprocessed
\end{enumerate}
In total 229 entries were downloaded, and repeated particle image were de-duplicated by the stack's basename and particle index.
Metadata is extracted from the MRCS or STAR files.
We include images without metadata (i.e., no CTF annotations), but cap the contribution to at most 200,000 particles per entry.
Unreadable stacks, invalid image dimensions, and out-of-range particle references were excluded.
Images are stored in float32 arrays in their native box sizes.
The final dataset contains 51,606,609 from 214 EMPIAR entries.

\paragraph{OpenProteinSet Processing.} We use the mmCIF files provided in the OpenProteinSet.
All obsolete PDB entries are excluded, and protein chains from the first structural model are extracted. 
We apply the biological assembly specific transforms specified in each mmCIF file, which contains separate metadata describing how to construct each assembly.
Each assembly is retained in our dataset.
Where assembly metadata was not available, the protein chains were concatenated.
Structures with resolutions \textgreater$5$\AA~were excluded, as well as assemblies with greater than 10,000 residues after symmetry expansion.
Structures with unknown resolution are kept.
Finally, the assembly coordinates are centered using the mean position of the atoms.
The resulting dataset contains 268,647 assemblies.

\paragraph{Pretraining Image Formation Arguments.} We use CryoJax for our image formation, specifically version $0.5.6$.
We simulate all atoms, using the Peng scattering factors~\citep{peng_robust_1996}.
For ViT-S/B/L, we render on a $640\times640$ grid and crop to $384\times384$ to reduce boundary artifacts from the CTF.
The higher-resolution ViT-g Gram-teacher branch is described in Appendix~\ref{appendix:augmentations}.
We randomly sample \AA/pixel values such that the rendered protein spans 35-80\% of the image extent.
We compute this projected span from the rotated atomic coordinates.
We list the parameters for the contrast transfer function in Table~\ref{tab:appendix-training-ctf}.
We use additive Gaussian white noise, with target SNRs sampled log-uniformly from $[0.1, 0.999]$.
The noise variance is the variance of the noiseless, CTF-applied image within a centered circular mask of radius half the image width, divided by the target SNR.

\begin{table}[htbp!]
\centering
\caption{CTF and envelope parameters used for synthetic pre-training.}
\label{tab:appendix-training-ctf}
\begin{tabular}{lll}
\hline
Parameter & Value or range & Sampling \\
\hline
Accelerating voltage & $300\,\mathrm{kV}$ & Fixed \\
Spherical aberration ($C_s$) & $2.7\,\mathrm{mm}$ & Fixed \\
Amplitude contrast ratio & $0.1$ & Fixed \\
Mean defocus & $0.9$--$2.5\,\mu\mathrm{m}$ & Uniform \\
Astigmatism & $0$--$200\,\mathrm{\text{\AA}}$ & Uniform \\
Astigmatism angle & $0^\circ$--$180^\circ$ & Uniform \\
Envelope B-factor & $10$--$150\,\mathrm{\text{\AA}}^2$ & Uniform \\
\hline
\end{tabular}
\end{table}

\section{Training Details}
\label{appendix:training-details}

\subsection{Model Losses}

\begin{figure}
    \centering
    \includegraphics[width=0.9\linewidth]{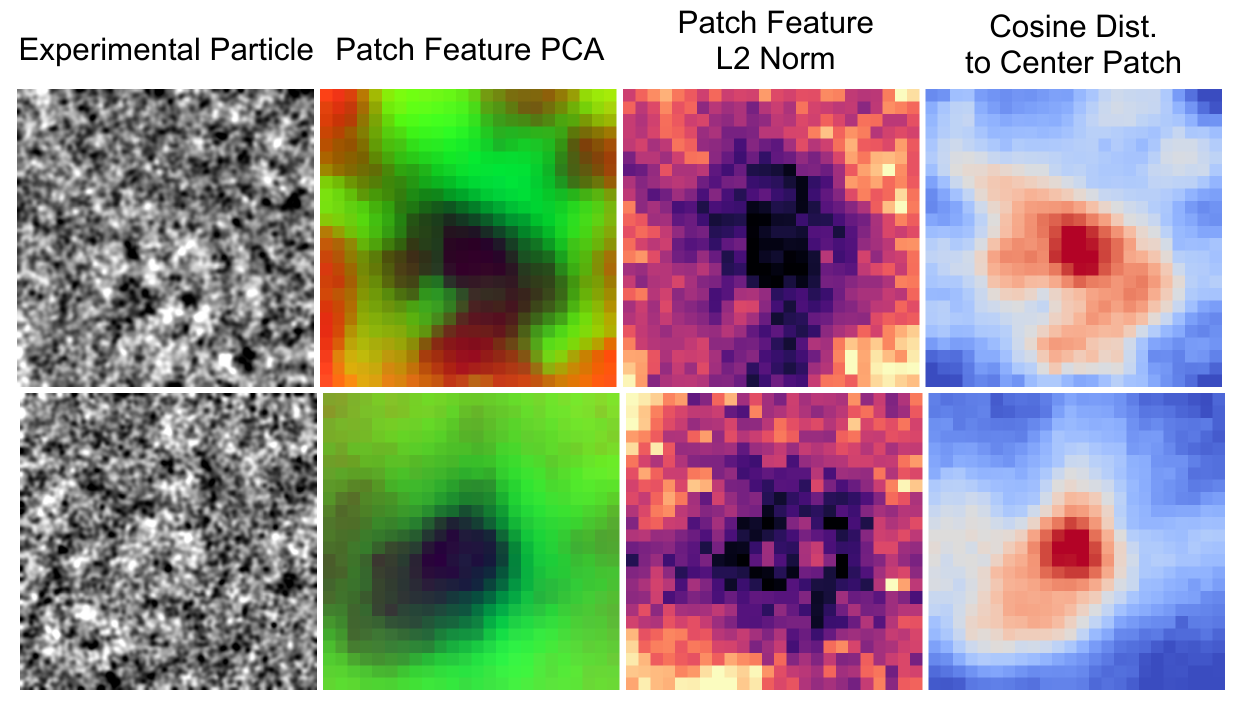}
    \caption{\textbf{Example Dense Features.} Experimental cryo-EM image, low-pass filtered for visibility (left), followed by the image features colored by PC1, PC2, and PC3, L2 vector norms of each patch, and, finally, the cosine distance between each patch's features and those of the center patch. \model~produces dense, semantically meaningful features.}
    \label{fig:dense-features}
\end{figure}

We recount the training losses from DINOv2 and DINOv3 here for completeness~\citep{oquab_dinov2_2024, simeoni_dinov3_2025}:

\paragraph{(1) DINO Image-Level Objective~\citep{caron_emerging_2021}.}
Both the teacher class token $v_t^{CLS}$ and the student class token $v_s^{CLS}$ are passed through an MLP that outputs a set of prototype scores. Student scores are normalized using a softmax with temperature $0.1$ to obtain $p_s$.
We use Sinkhorn-Knopp normalization to turn teacher scores into balanced target distributions.
The DINO image-level objective is then:
\begin{align}
    \label{eq:dino-loss}
    \mathcal{L}_{\mathrm{DINO}} = - \Sigma p_t \log p_s
\end{align}
Like the backbone, the student head is optimized using gradient descent, while the teacher head is an EMA of the student head.

We compare each student global view with the other teacher global view, excluding same-view pairs, and compare each of the eight student local views with both teacher global views.
This gives two global-to-global and sixteen local-to-global pairs per training item.
The loss sums over all prototypes, and averages over training items and these eighteen view pairs.

\paragraph{(2) Masked Image Modeling Objective~\citep{zhou_ibot_2021}.}
Certain patch tokens are masked when passed to the student network.
% note the dimensionality of the bottleneck is slightly different but this okay for this draft - we can fix it in the final conference paper
These patch tokens pass through the iBOT head, which is structured identically to the DINO head and produces normalized prototype scores $p_{si}$.
The corresponding tokens are unmasked when provided to the teacher, whose iBOT head produces $p_{ti}$. The iBOT masked image modeling objective is
\begin{align}
    \label{eq:ibot-loss}
    \mathcal{L}_{\mathrm{iBOT}} = -\Sigma p_{ti}\log{p_{si}}
\end{align}

We match teacher and student targets to the same patch positions within the same global view.
For each of the student's global view images, we average the cross entropy over its masked patches and then average over all global view images in the batch.
Finally, images without masked patches contribute nothing to the loss but are included in the denominator for the average.

\paragraph{(3) KoLeo Loss~\citep{sablayrolles2018spreading}}. The KoLeo loss term acts as a regularizer which promotes features to uniformly span the feature space. 
For feature vectors $(x_1, \ldots, x_n)$, it is defined:
\begin{align}
    \label{eq:koleo-loss}
    \mathcal{L}_{\mathrm{KoLeo}} = - \frac{1}{n}\sum_{i=1}^n \log(d_{n,i})
\end{align}
where $d_{n,i} = \min_{i\neq j}||x_i - x_j||$, the minimum distance from $x_i$ to any other feature vector. 
Feature vectors are $L_2$ normalized before computing the KoLeo loss.

\paragraph{(4) Gram Anchoring Loss.} We use the Gram anchoring loss from DINOv3 during our ViT-g training run.
When using the Gram anchoring, we keep a separate Gram teacher, from an earlier checkpoint, in memory which is used to compute the anchoring loss.
We enable this loss at step $187{,}500$, with a $1{,}000$-step cosine warmup to weight $2.0$.
The Gram teacher is frozen between updates and is refreshed from the EMA teacher after $197{,}500$, $207{,}500$, and $217{,}500$ completed optimizer steps.
Adopting the notation from DINOv3, suppose we have $P$ patch features of dimension $d$.
Let $\mathbf{X}_S$, a $P\times d$ matrix, denote the student dense features and $\mathbf{X}_G$ the dense features of the Gram teacher.
The Gram anchoring loss is:
\begin{align}
    \label{eq:gram-anchor-loss}
    \mathcal{L}_{\mathrm{Gram}}
    = \frac{1}{P^2}
      \sum_{i=1}^{P}\sum_{j=1}^{P}
      \left[
      (\mathbf{X}_S\mathbf{X}_S^\top)_{ij}
      -
      (\mathbf{X}_G\mathbf{X}_G^\top)_{ij}
      \right]^2.
\end{align}
Each patch-feature vector is $L_2$ normalized before constructing the similarity matrices.
We average this loss over global-view images.

\paragraph{Combined Loss Function.}
Put together, the pretraining loss function is:
\begin{align}
    \mathcal{L} = w_D \mathcal{L}_{\text{DINO}} +  \mathcal{L}_{\text{iBOT}} + w_K \mathcal{L}_{\text{KoLeo}} + w_G \mathcal{L}_{\text{Gram}}
\end{align}
We use a DINO loss weight of $w_D = 1.0$ and KoLeo loss weight of $w_K = 0.1$.
$w_G$ is set to 0 when Gram anchoring is disabled and reaches $2.0$ after the warmup when it is enabled.

When we use gradient accumulation, each microbatch loss is divided by the number of accumulation steps before backpropagation.

\subsection{Architectures, Optimizer, LR Schedule, Data-Mixing}
We inherit the architectural choices of DINOv3, and list the architectural hyperparameters of each model size in Table~\ref{tab:appendix-model-architectures}.
Hyperparameters for DINOv2 and DINOv3 are from Table 2 by~\cite{simeoni_dinov3_2025}.

\paragraph{Target Distributions, Temperatures, and Teacher Momentum.}
We use a softmax temperature of 0.1 for both student heads.
For the DINO teacher, the temperature is held constant at 0.04 for the first 1,250 optimizer steps, then increases linearly from 0.04 to 0.07 until step 30,000, at which point it remains constant for the duration of training.
We have a similar schedule for the iBOT teacher. Its temperature increases linearly from 0.04 to 0.07 from steps 0 to 30,000, and then remains constant.
For both teacher heads, we exponentiate the scores divided by their respective temperatures and apply three Sinkhorn-Knopp iterations.
This normalization helps to balance the prototype mass across all distributed ranks and produces a probability distribution over prototypes for each token.
DINO balancing uses all the teacher global view class tokens, while iBOT balancing uses only masked patches in the corresponding student inputs.
Unlike some DINO implementations, we do not use running mean-based teacher centering.
We update the teacher parameters using an exponential moving average with momentum 0.994, increased to 0.999 when we start using Gram anchoring.

\begin{table*}[t]
\centering
\caption{
Hyperparameter comparison between our models and the published DINOv2/v3.
}
\label{tab:appendix-model-architectures}
\begingroup
\small
\setlength{\tabcolsep}{4pt}
\renewcommand{\arraystretch}{1.12}
\resizebox{\textwidth}{!}{%
\begin{tabular}{@{}lcccccc@{}}
\toprule
& \multicolumn{4}{c}{\model}
& \multicolumn{2}{c}{Reference models} \\
\cmidrule(lr){2-5}
\cmidrule(l){6-7}
Property
& Small & Base & Large & Giant & DINOv2 & DINOv3 \\
\midrule
Backbone
& ViT-S/16 & ViT-B/16 & ViT-L/16 & ViT-g/16
& ViT-g/14 & ViT-7B/16 \\
Backbone parameters
& 21.4M & 85.3M & 304.2M & 1133.7M & 1.1B & 6.7B \\
Backbone + heads parameters
& 233.2M & 301.8M & 523.9M & 1359.7M & --- & --- \\
Transformer blocks
& 12 & 12 & 24 & 40 & 40 & 40 \\
Input channels
& 1 & 1 & 1 & 1 & 3 & 3 \\
Patch size (pixels)
& $16 \times 16$ & $16 \times 16$ & $16 \times 16$
& $16 \times 16$ & $14 \times 14$ & $16 \times 16$ \\
Positional encoding
& RoPE & RoPE & RoPE & RoPE & Learnable & RoPE \\
Register tokens
& 4 & 4 & 4 & 4 & 4 & 4 \\
Embedding dimension
& 384 & 768 & 1024 & 1536 & 1536 & 4096 \\
FFN type
& SwiGLU & SwiGLU & SwiGLU & SwiGLU & SwiGLU & SwiGLU \\
FFN hidden dimension
& 1024 & 2048 & 2752 & 4096 & 4096 & 8192 \\
Attention heads
& 6 & 12 & 16 & 24 & 24 & 32 \\
Attention head dimension
& 64 & 64 & 64 & 64 & 64 & 128 \\
Student drop-path rate
& 0.2 & 0.4 & 0.4 & 0.4 & 0.4 & 0.4 \\
LayerScale initialization
& $10^{-5}$ & $10^{-5}$ & $10^{-5}$
& $10^{-5}$ & $10^{-5}$ & $10^{-5}$ \\
\midrule
DINO head MLP
& 8192--8192--512
& 8192--8192--512
& 8192--8192--512
& 8192--8192--512
& 4096--4096--256
& 8192--8192--512 \\
DINO prototypes
& 131\,072 & 131\,072 & 131\,072
& 131\,072 & 131\,072 & 262\,144 \\
iBOT head MLP
& 4096--4096--384
& 4096--4096--384
& 4096--4096--384
& 4096--4096--384
& 4096--4096--256
& 8192--8192--384 \\
iBOT prototypes
& 131\,072 & 131\,072 & 131\,072
& 131\,072 & 131\,072 & 98\,304 \\
\bottomrule
\end{tabular}%
}
% \par\smallskip
% \begin{minipage}{\textwidth}
% \footnotesize

% \end{minipage}
\endgroup
\end{table*}

\paragraph{Optimization Details.} 
Across all models, we use a global batch size of 3072.
Depending on the available compute capacity, we use either no gradient accumulation or 2 steps.
For backbone parameters that are exactly two-dimensional tensors, we use Muon~\citep{jordan_muon_2024, liu_muon_2025} while remaining backbone parameters use AdamW.
Both iBOT and DINO head parameters are updated with AdamW~\citep{loshchilov_decoupled_2019}.
Muon uses a momentum of 0.95, Nesterov updates, and five Newton-Schulz iterations. We scale matrix updates by a factor proportional to the layer dimensions so that the update norm aligns with AdamW.
For AdamW, we use $\epsilon=10^{-8}$, with $(\beta_1,\beta_2)=(0.9,0.95)$ for the backbone and $(0.9,0.99)$ for both projection heads.
We use a weight decay of 0.05, excluding biases, normalization parameters, and LayerScale coefficients.

This choice follows empirical recommendations to optimize final prediction layers with AdamW when using Muon~\citep{jordan_muon_2024}; we extend this choice to the complete projection heads for a uniform head optimization scheme.
We use a base learning rate of $1\times10^{-4}$ tuned at a batch size of 256, scaling the learning rate by the square root of the target batch size (3072).
All models are trained for 265,000 optimizer steps with gradient clipping set to 3.0.
We use a warmup-stable-decay (WSD)~\citep{wen_understanding_2024} learning rate scheduler.
The global learning rate is linearly warmed up for 15,000 steps, held stable for 223,500 steps, and decayed using 26,500 steps (10\% of the total schedule).
Patch-embedding parameters use 0.2 times the global learning rate throughout training.
We freeze the final prototype projections of both the DINO and iBOT heads for the first 1,250 optimizer steps by setting their learning rates to zero, and linearly increasing to one over the following 13,750 steps, applied multiplicatively to the global learning-rate schedule. 
When running evaluations, we periodically branch and decay the learning rate over a 10\% cooldown period following $\eta(p) = \eta_{\mathrm{peak}}(1-\sqrt{p})$, where $p \in [0,1]$ denotes progress through the decay phase.

\begin{figure}
    \centering
    \includegraphics[width=0.8\linewidth]{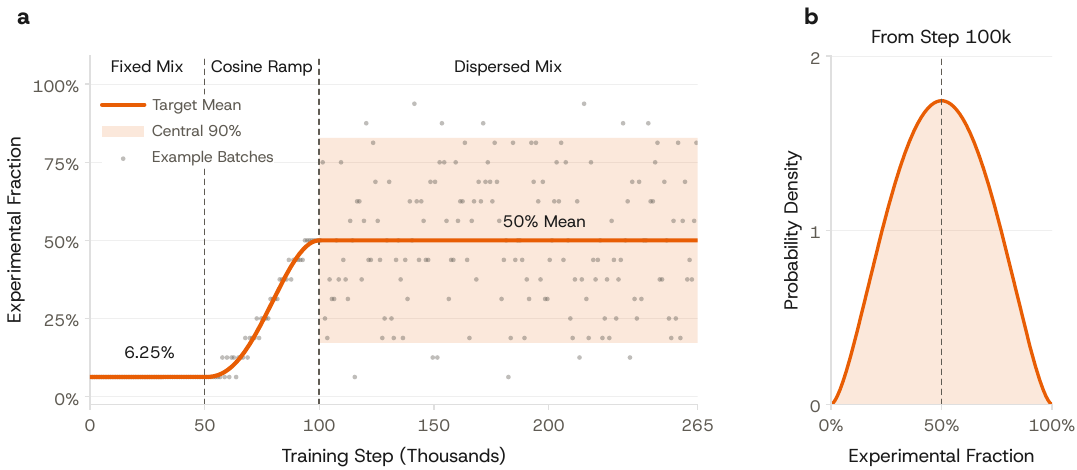}
    \caption{\textbf{Experimental data schedule.} Visualization of the proportion of training batch that comes from experimental data. We pre-train on primarily synthetic images, before using a cosine schedule to mid-training where we sample using a dispersion equation.}
    \label{fig:appendix-data-mixing}
\end{figure}

\paragraph{Data-Mixing Schedule.} 
We find that introducing too much experimental data at the start of training destabilizes training and causes representation collapse.
As a result, we use a heuristic schedule which we find to stably train.
During the first 50,000 steps, as part of pre-training, experimental images comprised 6.25\% of each batch. 
We transition to mid-training by increasing the experimental data mix to 50\% over the next 50,000 steps using a cosine ramp with a progress exponent of 1.2.
From step 100,000 onward, we maintained a mean experimental fraction of 50\% and vary the batch composition by using a beta distribution with a dispersion factor of 0.4: $f_{\text{exp}} \sim \text{Beta}(2.625, 2.625)$.
\Fref{fig:appendix-data-mixing} shows the continuous schedule. 
In practice, we round the experimental fraction to adjacent multiples of 6.25\% to limit the number of batch shapes requiring compilation on each rank in CryoJax.

\subsection{Augmentations.}
\label{appendix:augmentations}

For simulated data, we generate different views by varying the orientation and imaging parameters in CryoJax.
For experimental data, we instead augment recorded particle images using the pairing and image transformations described below.
We follow DINOv3~\citep{simeoni_dinov3_2025} in using two global views and eight local views per training item, with resolutions of $384\times384$ and $128\times128$, respectively, in our setup.

\paragraph{Synthetic pairs.}
We render paired views of the exact same molecular assembly and conformation, keeping the underlying 3D structure fixed.
We first sample an orientation uniformly over 3D rotations, then with probability $p=0.5$ independently sample a second orientation; otherwise, we reuse the first.
For each source projection, we independently choose the pixel size so that the projected molecular span occupies a uniformly sampled $35$--$80\%$ of the image width.
We independently sample CTF parameters, translations, noise levels, and noise realizations for each rendered view, including when the two source projections share an orientation.
Translations are sampled uniformly within $\pm15\%$ of the image width on each axis, and the CTF ranges are given in Table~\ref{tab:appendix-training-ctf}.
We render four additional images from each source projection to form the eight local crops.

\paragraph{Experimental pairs.}
We augment experimental particles with random translations and in-plane rotations.
With probability $p=0.5$, we pair a particle with another particle from the same EMPIAR entry; otherwise, both source images use the same particle.
The paired particles may differ in orientation, conformation, or composition; a shared EMPIAR entry does not guarantee the same underlying 3D structure.
For ViT-S/B/L, we resize each source image to $461\times461$ using bilinear interpolation, then sample $384\times384$ windows to produce translations.
For each window, we independently sample the horizontal and vertical top-left coordinates uniformly from the integers $0$ through $77$.
This keeps the window within the resized image without adding boundary pixels.
The global views use one window from each source image, and we sample four additional windows from each image for the local crops.
After extracting the translation windows, we independently rotate each global window and each window used for local cropping by an angle sampled uniformly from $\{0^\circ,90^\circ,180^\circ,270^\circ\}$.
These angles are newly sampled each time the particle is loaded.

\paragraph{Local crops and normalization.}
We use per-image standardization, subtracting the pixel mean of each global image and each image used for local cropping, then divide by its sample standard deviation, clamped below at $10^{-6}$.
We retain DINOv3's random resized cropping procedure, but set the local crop area range to $25\%$--$50\%$ of the image area.
The area is sampled uniformly, and the aspect ratio is sampled log-uniformly between $3/4$ and $4/3$.
The crop is resized to $128\times128$ using bicubic interpolation with anti-aliasing.
In our implementation, we use the same crop window for all local images within each per-device batch, and do not standardize again after cropping.

\paragraph{Teacher and student inputs.}
Following DINOv3, the EMA teacher receives the two unmasked global views, while the student receives both global views and all eight local views.
For the iBOT loss, we retain DINOv3's masking settings: we replace patch embeddings with a learned mask token in half of the student global views; the local views remain unmasked.
Target masking ratios are evenly spaced within $10\%$--$50\%$ across the masked views, and we shuffle the mask assignment across the global views in the batch.
We form block masks and fill any remaining positions by randomly selecting unmasked patches until the target count is reached.

\paragraph{Gram-teacher views.}
Before Gram anchoring is enabled, ViT-g uses the same view-generation procedure as ViT-S/B/L.
Following DINOv3's high-resolution Gram anchoring procedure~\citep[Sec.~4.3]{simeoni_dinov3_2025}, the ViT-g Gram teacher receives global views at twice the student resolution.
In our setup, these are two $768\times768$ views, which we standardize separately using the same per-image rule.
For synthetic data, this branch uses $1152\times1152$ padding; the ordinary $384\times384$ views use Fourier-downsampled projections with $576\times576$ padding.
Corresponding high- and low-resolution views share the orientation, field of view, CTF parameters, translation, and target SNR, but use independent noise realizations.
For experimental data, we instead resize images to $922\times922$ and sample $768\times768$ windows, with top-left coordinates from $0$ through $154$ on each axis.
We apply the same rotation procedure, retain the two global views at this resolution for the Gram teacher, and downsample all windows to $384\times384$ before forming the ordinary global and local views.
Before computing the Gram loss, we resize the teacher's $48\times48$ patch-feature grid to the student's $24\times24$ grid using bicubic interpolation with anti-aliasing and corner alignment disabled.
We compute this loss over all spatial patches, including those masked in the student.

\section{Evaluation details}
\label{appendix:evaluations}

\paragraph{Computing FLOPs.}
We report floating point operations (FLOPs) to estimate the computation needed for one forward pass through the encoder on a single image.
Specifically, for a ViT with $P$ backbone parameters, image side length $S$, patch side length $p$, and $R$ register tokens, we use
\begin{equation}
    F_{\mathrm{backbone}} \simeq 2 P N_{\mathrm{tok}},
    \qquad N_{\mathrm{tok}} = (S/p)^2 + 1 + R.
    \label{eq:backbone-flops}
\end{equation}
We count a multiplication and an addition as separate operations, giving the factor of two, and include the CLS token in the token count.
Unless otherwise specified, we use an image size of $384^2$, which yields different token counts based on the model's patch size.
\model{} uses $p=16, R=4$, Cryo-IEF uses $p=14, R=0$, and CryoUNI uses $p=16, R=4$.
We compute FLOPs separately for each encoder, rather than assigning the baselines the same FLOPs as the corresponding \model{} size.

\paragraph{Image preprocessing.}
We use different preprocessing for each model, reproducing publicly documented inference pipelines when possible.
For \model{}, we standardize each image to zero mean and unit variance.
For Cryo-IEF, we first apply min--max scaling to each image, and then normalize using the mean and standard deviation in their GitHub repo.
For CryoUNI we standardize each image to zero mean and unit variance, CTF phase flip (when known), and use Fourier-resampling if applicable.

\paragraph{Model checkpoints.}
For comparisons to Cryo-IEF and CryoUNI, we use their publicly available checkpoints, through HuggingFace for Cryo-IEF and Zenodo for CryoUNI.

\begin{itemize}
    \item Cryo-IEF: \url{https://huggingface.co/westlake-repl/Cryo-IEF},\newline Commit Hash: c259a597febe320fad5b9dade1fc98dfeea6b777
    \item CryoUNI Zenodo DOI: \verb|10.5281/zenodo.19513129|
\end{itemize}

\subsection{CryoPPP Composition Heterogeneity kNN Classification}
\label{appendix:cryoppp-knn}

\begin{figure}[htbp!]
    \centering
    \includegraphics[width=0.8\linewidth]{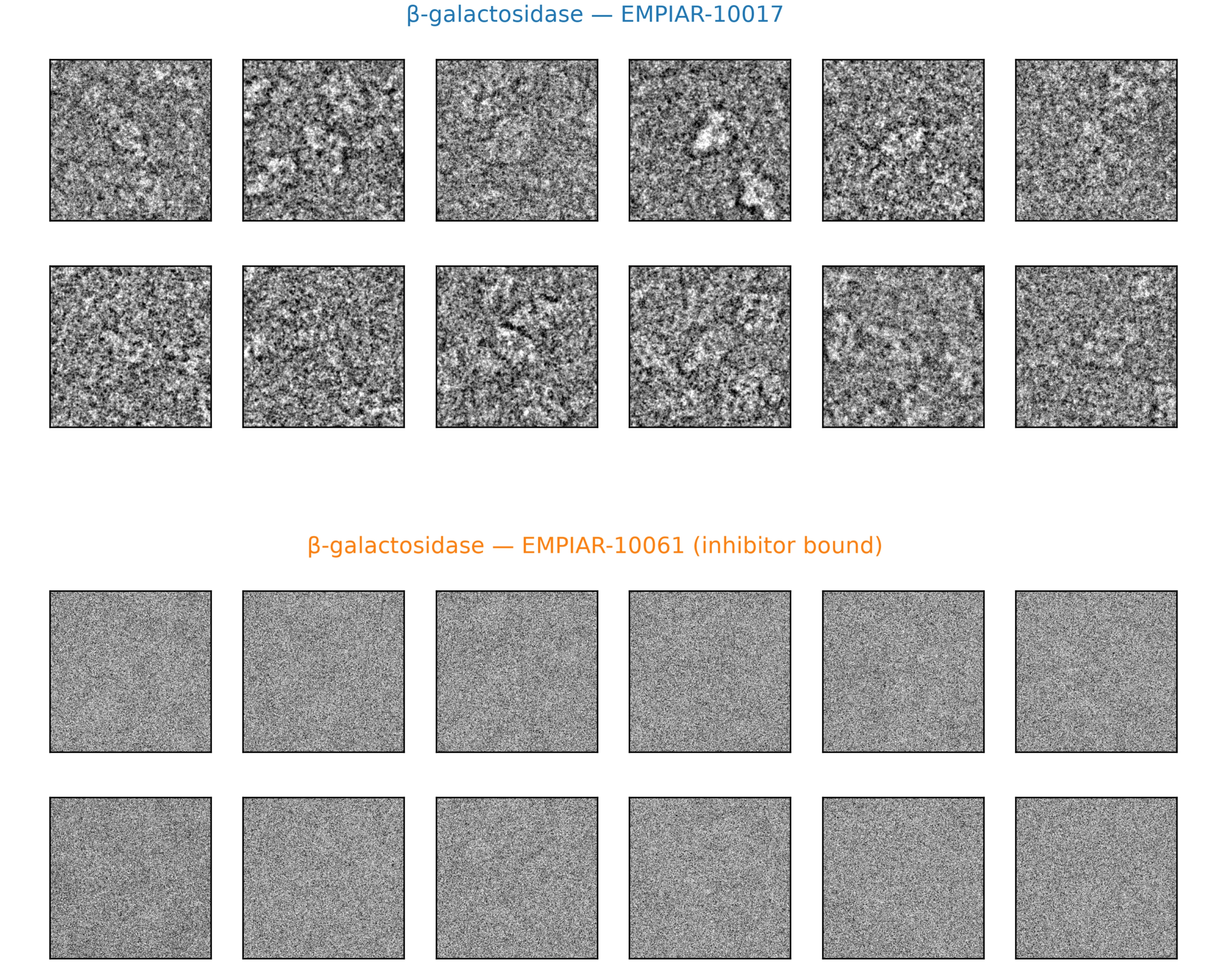}
    \caption{\textbf{Visualized CryoPPP Experimental Images.} (Top) EMPIAR 10017 $\beta$-galactosidase and (Bottom) EMPIAR 10061 also $\beta$-galactosidase but bound to an inhibitor. Structurally similar, but present visually different, potentially due to experimental conditions.}
    \label{fig:sample-cryoppp-images}
\end{figure}

\paragraph{Sampling.}
We use picked particles from 29 EMPIAR entries in CryoPPP ~\citep{dhakal_cryoppp_2023}.  We start with the ground-truth selected-particle lists, then retain entries with at least 1,250 usable particles and sample this many from each entry without replacement using a fixed seed.  We assign 1,000 particles per entry to the labeled feature bank and the remainder to the query set, so in the end we have 29,000 labeled feature images and 7,250 query images.

\paragraph{EMPIAR IDs.}
In this evaluation, we use EMPIAR entries 10005, 10017, 10028, 10059, 10061, 10075, 10081, 10093, 10096, 10184, 10240, 10289, 10291, 10345, 10387, 10406, 10444, 10526, 10532, 10576, 10590, 10669, 10671, 10737, 10760, 10816, 11051, 11057, and 11183.

% A common method used in the natural vision community test the semantic information of representations is via a k-nearest neighbor probe using ImageNet.

\paragraph{Probe methodology.}
In this probe, a set of labeled training images are embedded using the model and a final representation is produced by concatenating the \verb|[CLS]| token and mean pooling the spatial patch features~\citep{oquab_dinov2_2024}.  For \model{}, we $\ell_2$-normalize each concatenated representation in both the labeled feature bank and the query set.  For each query, we select the $k$ bank particles with highest cosine similarity.  We assign each selected neighbor a weight proportional to $e^{s/0.07}$, where $s$ is the cosine similarity to the query, and normalize these weights over the selected neighbors.  We sum the weights by EMPIAR class and rank classes by their total weight to obtain the top-1 and top-5 predictions.  
Similar to the natural vision case, we use 29 EMPIAR datasets of picked particles from CryoPPP~\citep{dhakal_cryoppp_2023}, a dataset used for particle picking.  We sweep over $k=1,5,10,20$ and report the highest top-1 and top-5 accuracy separately.  In the case of a tie we choose the smaller $k$.  We use the same query set for selecting $k$ and determining accuracy.
% A similar evaluation is done in Cryo-IEF~\citep{yan_comprehensive_2026}.  
% \nc{similarity metric, feature normalization, and voting rule}

\begin{figure}[htbp]
    \centering
    \includegraphics[width=\linewidth]{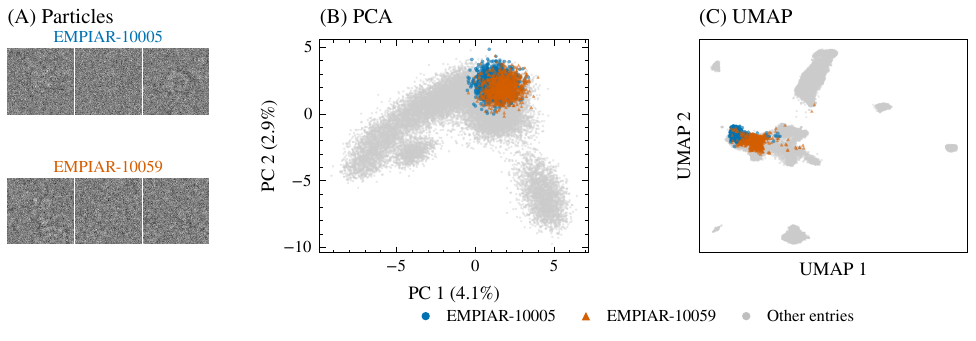}
    \caption{\textbf{Exploratory TRPV1 visualization.}
    (A) Three randomly sampled query particles from each entry,
    resized to $384\times384$ and standardized separately.
    Intensities are displayed over $[-3,3]$ standard deviations;
    this display clipping is not applied during evaluation.
    The images are not shown at a common physical scale.
    (B,C) PCA and UMAP fitted to normalized CLS features from
    all 29,000 feature-bank particles, using a ViT-S teacher
    checkpoint at step 79,999.
    Blue circles indicate EMPIAR-10005, orange triangles indicate
    EMPIAR-10059, and the other 27 entries are gray.
    UMAP uses 15 neighbors, a minimum distance of 0.1, and seed 42.
    This earlier checkpoint and CLS-only representation differ
    from those used for the reported $k$NN results.}
    \label{fig:cryoppp-trpv1}
\end{figure}

\paragraph{TRPV1 example.}
We visualize two entries containing TRPV1, EMPIAR-10005 and EMPIAR-10059, using PCA and UMAP (\Fref{fig:cryoppp-trpv1}).
The entries overlap substantially in PCA and show partial
separation in UMAP.
These projections do not distinguish molecular differences
from acquisition-specific effects.

\subsection{CAHRA Probing Settings}
\label{appendix:cahra-probe}

% important items
% \begin{itemize}
%     \item How we select data
%     \item how we control for head parameter counts
%     \item how we compute the visualization axis for the plot
%     \item include the probe head comparison across the different models
% \end{itemize}

\paragraph{Dataset.}
We sampled uniformly without replacement within each of six species (ADH, aldolase, $\beta$-amylase, GDH, IgG, and PKM2), and selected 10,000 particles per species for training (60,000 total) and another 10,000 per species from the remaining particles for held-out evaluation (60,000 total). 
CAHRA images are provided at a native $288\times288$ pixel resolution. However, prior work demonstrates that training at a lower resolution than test-time results in higher overall results~\citep{touvron2019fixing}. For this reason, and to provide an equal-FLOPs comparison, we use bilinear interpolation to resize images to $384^2$, zero-padding images when necessary to make image dimensions divisible by the patch size.
We further ensure that the training and validation sets do not overlap.
The experimental particles in the challenge dataset are not contained in the pre-training dataset either. 

\paragraph{Probing Details.}
When running the evaluation, we sweep over learning rates $3\times10^{-5}, 1\times10^{-4}, 3\times10^{-4}, 1\times10^{-3}, 3\times10^{-3}$ using AdamW (weight decay$=0.05$), use a batch size of 256, and take the average over three seeds.
We train for 100 epochs, using a learning rate warmup of 5 epochs and a cosine decay for 95 epochs, ending at min learning rate $10\%$ of the start. 
When linear probing, we use a single linear layer which takes in the \verb|[CLS]| token concatenated with mean pooled patch features.
The MLP head is structured to be a linear head, LayerNorm, GELU, and a final linear layer.
The input is similarly the concatenated \verb|[CLS]| and mean pooled patch features.
The CNN is a 3x3 convolution, LayerNorm over channels, GELU, global average pooling, and a final linear layer.
We use a hidden feature dimension of 128 for the convolutions, while the MLP hidden dimensions are selected to match the parameter count of the CNN head.
We report results over 3 seeds.

\paragraph{Fine-tuning Details.}
We use the same 60,000 training images and 60,000 validation images when doing fine-tuning. 
We add a linear classification head to the backbone. The linear classification head receives the concatenation of the CLS token and mean-pooled spatial tokens. The entire model is then optimized with a cross entropy loss.
We train for 20 epochs using AdamW (weight decay$=0.05$, $\beta_1=0.9, \beta_2=0.999$), and batch size of 1,024. We sweep learning rates of $1\times10^{-5}, 3\times10^{-5}, 1\times10^{-4}, 3\times10^{-4}, 1\times10^{-3}$, using one epoch of linear warmup followed by cosine decay.% to 10\% of the peak learning rate. Each configuration used a single random seed, 43. 
% Training used BF16 mixed precision, activation checkpointing, and gradient clipping at norm 1.0. Validation was performed before training and every two epochs; the learning rate
%   and checkpoint were selected by maximum validation top-1 accuracy among completed runs. Our models used 384-pixel inputs; CryoIEF and CryoUNI were evaluated at their native input sizes of 224 and 128 pixels, respectively, with an additional full learning-rate sweep
%   for both baseline families at 384 pixels.
\subsection{Hsp90 Finetuning Conformational Heterogeneity}
\label{appendix:hsp-finetune}

\paragraph{Task Description.}
Hsp90 is a protein with a conformational change described by the opening of its two chains, CMA and CMB, which we discretize.
% CMA classes are one degree apart, and CMB classes are two degrees apart.
There are, in total, 400 conformational structures.
We simulate 200,000 training images across SNR values of $0.1\text{--}0.01$, sampling different defocus ($0.5\text{--}2.0$ $\mu\mathrm{m}$), b-factors ({1, 100} $\AA^2$), and use an amplitude contrast of $0.1$.
The images are simulated at a $256^2$ box and an $A_{pix}$ of $1.0 \AA$, and center-cropped to $128^2$.
During simulation, we apply a random translation sampled from $[0, 20]$\AA.
For the final evaluation, we upsample the images from $128^2$ to $384^2$, our final training resolution.
% We intentionally do no re-simulate, as this forces a natural Nyquist frequency that can be reached.
Finally, we generate three different evaluation sets, corresponding to different SNR levels of $0.1, 0.01,$ and $0.001$.

\paragraph{Optimization setting.}
We do full fine-tuning given the training dataset and hold the overall optimization arguments the same across models and checkpoints. 
In particular, we use Adam (weight decay = $0$, $\beta_1 = 0.9, \beta_2 = 0.999)$, batch size of 64, 10,000 optimizer steps.
We sweep over learning rates $ 1e^{-5}, 3e^{-5}, 1e^{-4}, 3e^{-4}, 1e^{-3}$.
The learning rate is warmed up over 1,000 steps to the peak learning rate and then annealed to zero using a cosine schedule over the remaining 9,000 steps.
The fully-supervised ResNet-18 baseline uses 50,000 optimizer steps, and is tuned using the same learning rates as above.

\begin{figure}[htbp!]
    \centering
    \includegraphics[width=1.0\linewidth]{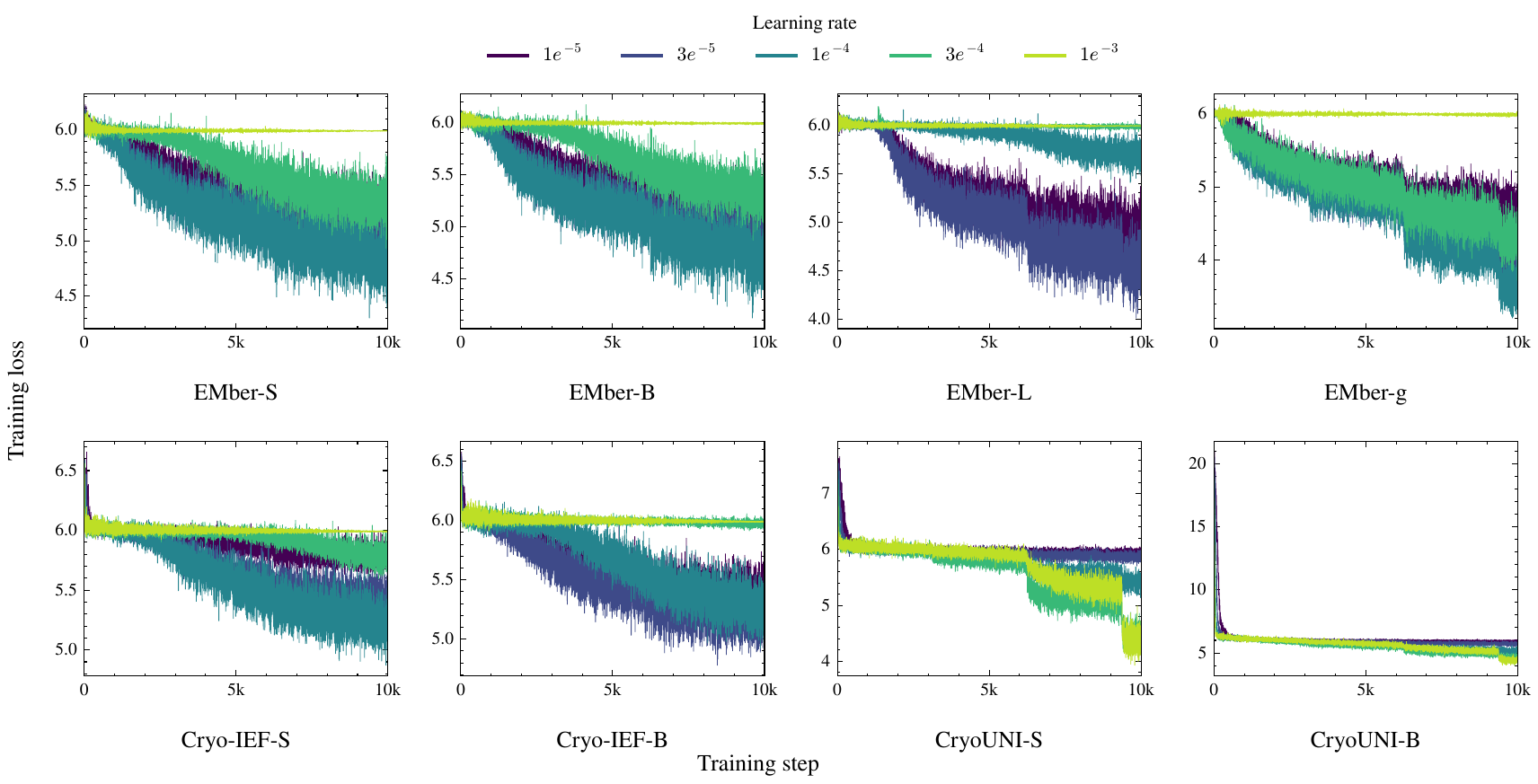}
    \caption{\textbf{Hsp90 Fine-tuning loss curves.} We visualize the training loss across different learning rates when fine-tuning on the Hsp90 dataset.}
    \label{fig:placeholder}
\end{figure}

% \subsubsection{Large Image Budget Fine-tuning}
% \nc{todo: results on if we train with 1M images, and for a longer horizon}

\subsection{Synthetic Spike Population Estimation}
\label{appendix:spike-pop-est}

\begin{figure}[tbph!]
    \centering
    \includegraphics[width=\linewidth]{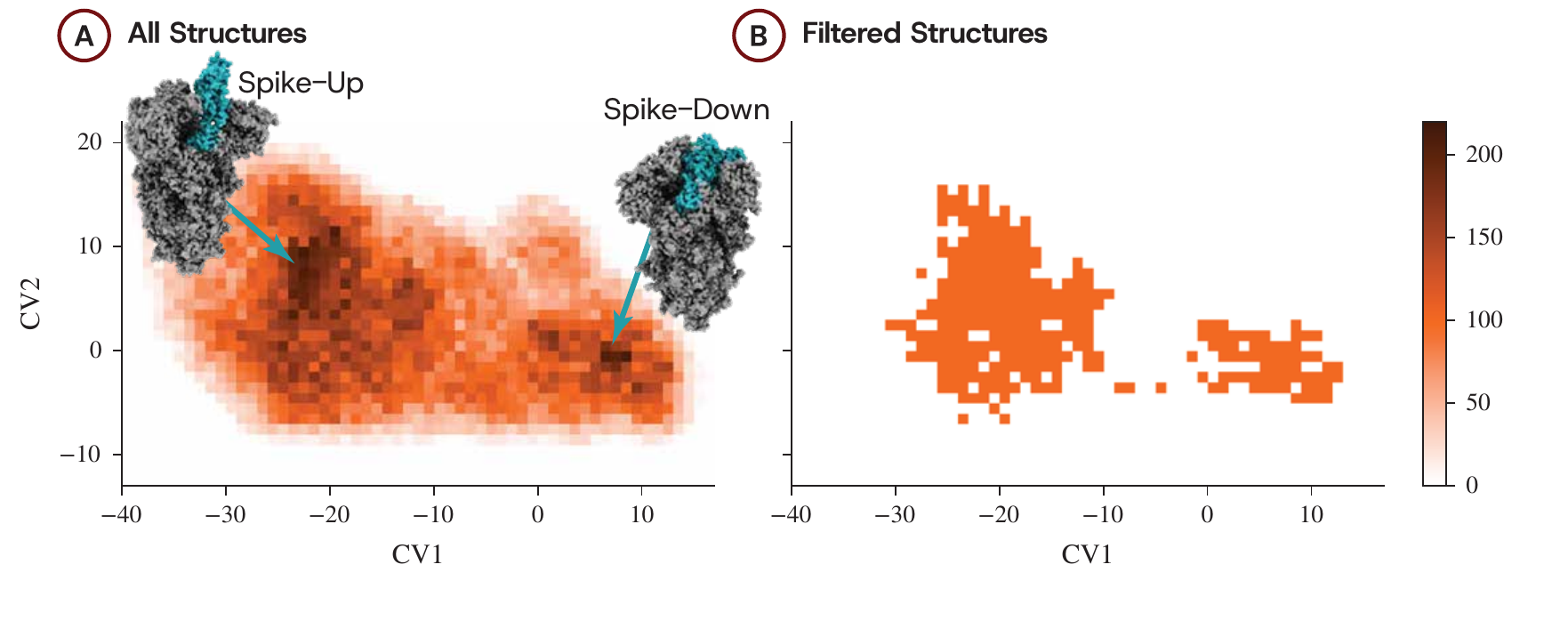}
    \caption{Spike protein structures from the Spike MD simulation used for Spike population estimation task plotted as a function of the collective variables (principal component 1, principal component 2).
    The original distribution of structures is shown in (A), with the filtered structures in (B).
    }
    \label{fig:spike-cv-filtering}
\end{figure}

\begin{figure}[tbph!]
    \centering
    \includegraphics[width=0.9\linewidth]{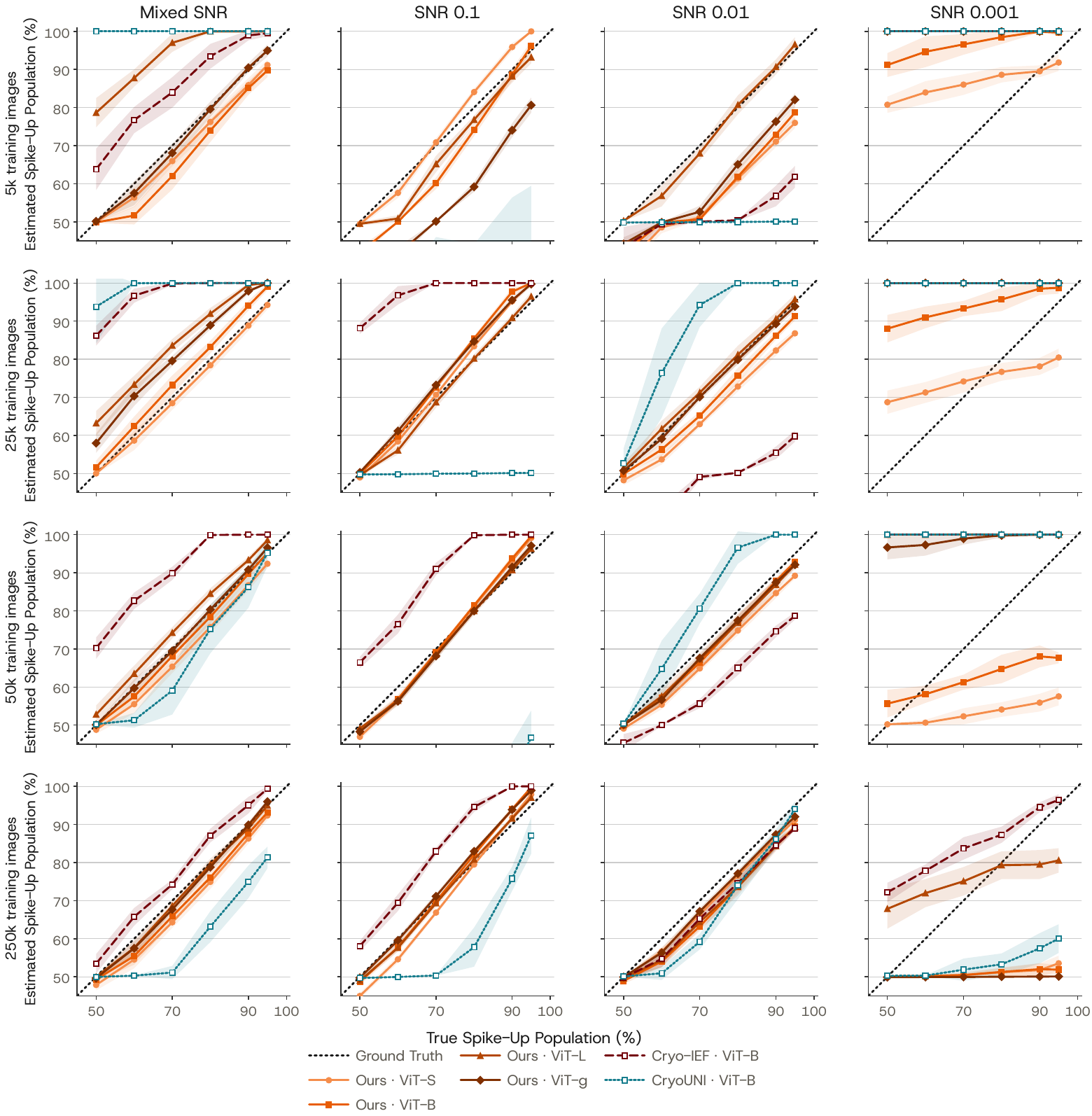}
    \caption{True Spike-Up population vs predicted Spike-Up population across a range of training dataset sizes used to fit the linear classifier, and stratified based on evaluation SNR level.}
    \label{fig:spike-pop-est-all}
\end{figure}

\paragraph{Population estimation and uncertainty.}
We treat the Spike conformations as two stable states, Spike-Up ($U$) and Spike-Down ($D$), each containing multiple underlying structures.
Let $\pi=p(U)$ denote the Spike-Up population, so that $1-\pi=p(D)$.
Given a set of particle images $\{y_i\}_{i=1}^N$ sampled from an unknown mixture, our goal is to estimate the underlying population $\pi_{\mathrm{true}}$.

For each image, we compute a representation using the frozen encoder and apply the trained binary linear probe.
If $\ell_{U,i}$ and $\ell_{D,i}$ are the two pre-softmax outputs, we define the scalar probe score as $s_i=\ell_{U,i}-\ell_{D,i}$, so that the uncalibrated Up probability is $q_i=\sigma(s_i)$.

Because a discriminative probability $p(U\mid y_i)$ is not itself a class-conditional likelihood, we calibrate the probe scores using an independent validation set containing equal numbers of Up and Down particles.
Specifically, we fit a logistic calibration model $p(U\mid s)=\sigma(as+b)$.
Since the calibration set has equal class priors, Bayes' rule implies that the posterior odds equal the class-conditional likelihood ratio.
So, we use $\hat{\lambda}_i=as_i+b$ as an estimate of the score-space log-likelihood ratio $\log[p(s_i\mid U)/p(s_i\mid D)]$, and define $\hat r_i=\exp(\hat{\lambda}_i)$.

This assumes that the class-conditional score distributions $p(s\mid U)$ and $p(s\mid D)$ remain fixed as the population mixture changes, such that different evaluation sets differ only in their Up/Down mixture weights.

Given a candidate population $\pi$, the score distribution is $p(s_i\mid\pi)=\pi p(s_i\mid U)+(1-\pi)p(s_i\mid D)$. Up to terms independent of $\pi$, the likelihood of the full dataset is

\begin{equation}
    L(\pi)
    \propto
    \prod_{i=1}^N
    \left[(1-\pi)+\pi\hat r_i\right].
    \label{eq:population_likelihood}
\end{equation}

We maximize Eq.~\ref{eq:population_likelihood} using expectation-maximization (EM). At iteration $t$, the responsibility that particle $i$ belongs to the Up state is $w_i^{(t)}=\pi^{(t)}\hat r_i/ [\pi^{(t)}\hat r_i+(1-\pi^{(t)})]$, and the population is updated as

\begin{equation}
    \pi^{(t+1)}
    =
    \frac{1}{N}\sum_{i=1}^N w_i^{(t)}.
    \label{eq:population_em}
\end{equation}

We initialize $\pi^{(0)}=0.5$ and iterate until convergence. For fixed per-particle likelihood ratios, this binary mixture-weight EM procedure maximizes Eq.~\ref{eq:population_likelihood} and returns the maximum-likelihood estimate of $\pi$.
The soft assignments $w_i^{(t)}$ are per-particle responsibilities used by EM and are distinct from uncertainty over the population parameter itself.
We report the mean population estimate across three probe-training seeds and ten population-sampling seeds, with shaded regions indicating ±1 standard deviation across these estimates.

% NC: this is not true, so I'm removing it
% To quantify population uncertainty, we retain the full likelihood over $\pi$ rather than only its maximizing value.
% For the primary analysis we use the uniform prior $\pi\sim\mathrm{Beta}(1,1)$, giving

% \begin{equation}
%     p(\pi\mid s_{1:N})
%     \propto
%     \prod_{i=1}^N
%     \left[(1-\pi)+\pi\hat r_i\right].
%     \label{eq:population_posterior}
% \end{equation}

% Because the $\mathrm{Beta}(1,1)$ prior is uniform, it does not shift the maximum. The MAP estimate coincides with the maximum likelihood estimate obtained by converged EM.
% We use it to retain the full posterior over $\pi$, allowing us to distinguish sharply identified population estimates from broad, weakly identified ones.
% We evaluate this posterior on a dense grid over $\pi\in[0,1]$ and report the posterior median and 95\% credible interval.

\paragraph{Structure Filtering.}
The molecular dynamics trajectories contain 46,789 frames of SARS-CoV-2 spike protein, which contain a continuum of open and closed states of the receptor binding domain.
This motion can be described with two collective variables (CV), visualized in~\Fref{fig:spike-cv-filtering}(A).
To filter and balance the set of structures we use, we bin the CV space using 1$^\circ$ increments.
We keep bins with at least 100 structures, and for bins with more than 100 structures we randomly sample 100 structures from the bin.
As a result, there are 4,500 structures for Spike-Up and another 4,500 for Spike-Down, for a total of 9,000 structures.

\paragraph{Image formation and Dataset.} 
Images were rendered at $384^2$ box size with a pixel size of 0.75 \AA.
Defocus is sampled uniformly from 0.5–-2.0 $\mu\mathrm{m}$, and B-factor is sampled uniformly from 1--100 $\AA^2$.
Amplitude contrast is fixed at 0.1.
The microscope voltage is simulated at 300kV.
The noise model used is Gaussian white noise, added after applying the CTF, and the variance is computed according to the expected SNR and by computing the variance of the CTF-applied image, with a circular mask of radius $192$.
Noise was applied across the entire image, and images were stored as unnormalized float16 arrays.

The dataset contained 400,000 images: 250,000 training images and three validation pools of 25,000 images each at fixed SNRs of 0.1, 0.01, and 0.001. 
We additionally have another 100,000 disjoint images used for the Platt calibration.
Training SNRs were sampled continuously and uniformly in log space over 0.001–0.1.

\paragraph{Optimization Settings.}
We train a linear classifier on the model features, mapping from the representation to a binary prediction task.
We train probes across a range of training images, including 5k, 25k, 50k, and 250k training images.
The images are drawn uniformly from the up and down conformations.
Similar to the Hsp90 setting, we stratify the validation sets according to SNR: $0.1, 0.01$, and $0.001$.
At each SNR level, we generate 25,000 images per class, evaluating ratios of $0.50, 0.60, 0.70, 0.80, 0.90$ and $0.95$.
For evaluation, each fixed-SNR validation pool was divided into 10,000 images for learning-rate selection and 15,000 images for final evaluation, with equal class counts in both partitions.
Population estimation used mixtures of 5,000 images sampled without replacement, across 10 sampling seeds. 
Mixed-SNR evaluation pooled the three final validation partitions into a 45,000-image pool before mixture sampling.
We use three different optimizer seeds when running the evaluation, training a new classifier for each seed.
This gives us 30 different population estimates per ground truth proportion.
We sweep over learning rates $1e^{-4}, 3e^{-4}, 1e^{-3}, 3e^{-3}, 1e^{-2}, 3e^{-2}, 1e^{-1}, 3e^{-1}, 1.0, 1.5$ for batch size 256 when training the probe, using the held out validation loss to select the optimal learning rate.
We train for 100 epochs, using a cosine annealing learning rate scheduler, and batch size 1024 with learning rate scaled according to the square root scaling rule.
The EM algorithm is capped to a max of 50k iterations, and we use a convergence tolerance of 0.01.

\subsection{Synthetic Structure Reconstruction}
\label{appendix:synthetic-reconstruction}

\begin{figure}[htbp!]
    \centering
    \includegraphics[width=0.8\linewidth]{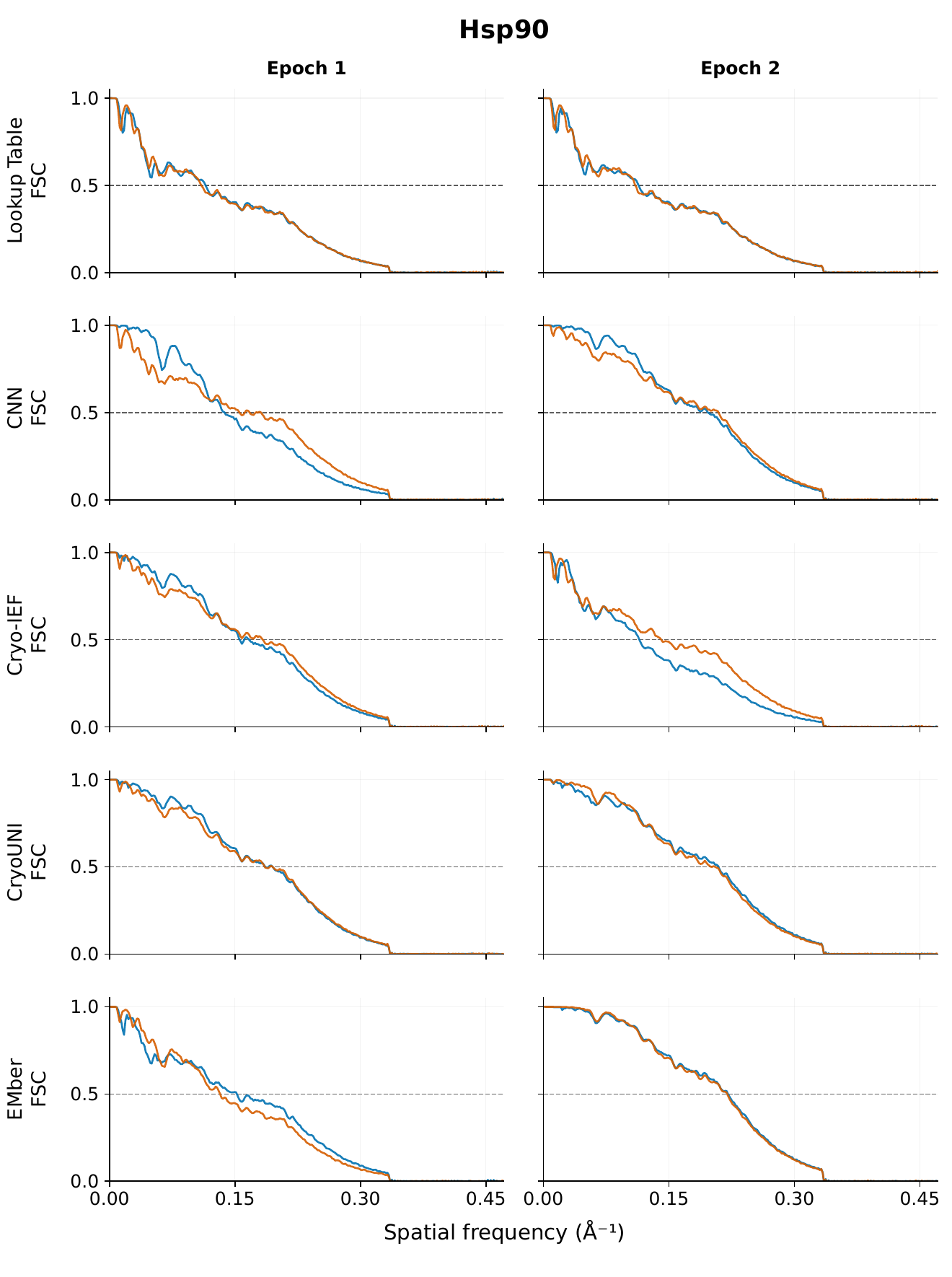}
    \caption{\textbf{Fourier Shell Correlation plots of Hsp90 reconstructions.} Rows correspond to different methods, and columns correspond to pose search epochs. The two curves represent the open and closed states. The dashed line indicates FSC=$0.5$.}
    \label{fig:appendix-fsc-hsp}
\end{figure}

\begin{figure}[htbp!]
    \centering
    \includegraphics[width=0.8\linewidth]{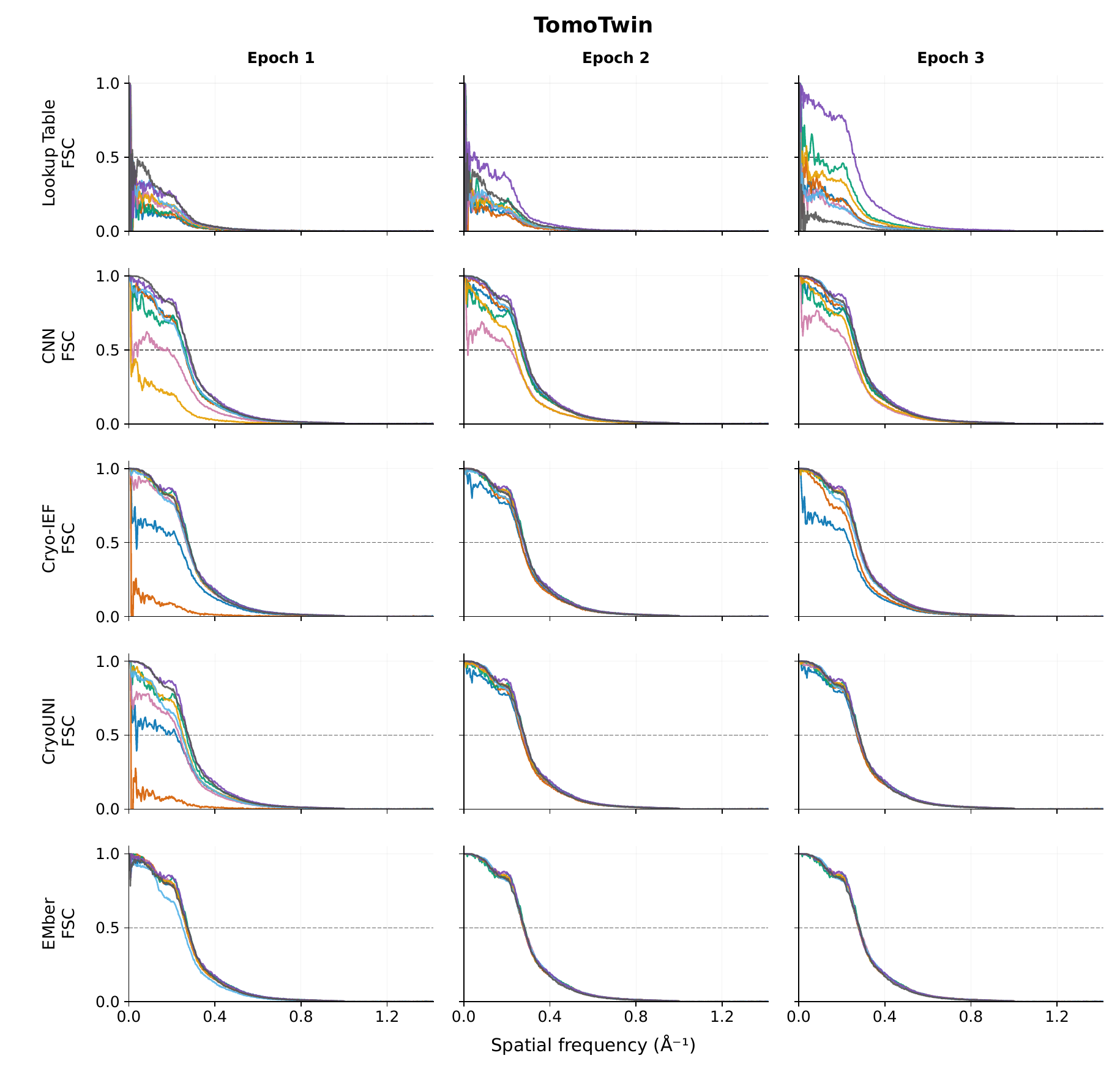}
    \caption{\textbf{Fourier Shell Correlation plots of Tomotwin-8 reconstructions.} Rows correspond to different methods, and columns correspond to pose search epochs. Each color represents one of the 8 reconstructed structures. The dashed line indicates FSC=$0.5$.}
    \label{fig:appendix-fsc-tomotwin}
\end{figure}

\paragraph{Evaluating Map Reconstructions.}
\label{appendix:resolution-explained}
Due to the nature of cryo-EM, there are no ground-truth labels for the structures.
The canonical method for determining map quality is the resolution derived from the Fourier shell correlation (FSC).
The FSC is the normalized cross-correlation between corresponding Fourier shells of two volumes.
When evaluating a map against a known reference, the spatial frequency at which the FSC crosses 0.5, corresponding to equal signal and noise, is reported as the resolution.
When a reference map is not available, the observed images are split into two half-sets that independently undergo reconstruction.
The spatial frequency at which the FSC crosses 0.143 is then reported as the resolution of the resulting map.
For a full derivation, refer to~\citet{rosenthal_optimal_2003}.
For both Tomotwin-8 and Hsp90, we visualize the FSC curves in~\Fref{fig:appendix-fsc-hsp} and~\Fref{fig:appendix-fsc-tomotwin}.

\paragraph{RELION Reconstruction Parameters and Resolution Assignment.} 
For each particle cluster, we reconstruct one volume with the ground truth poses using the \verb|relion_reconstruct| command in RELION version 4.0.1. Reconstructions used C1 symmetry, CTF correction, and the original full-resolution particle images with ground-truth orientations, translations, and CTF parameters from the source STAR files. 
Reconstruction pixel sizes were 1.5~\AA~for Hsp90 and 0.5~\AA~for TomoTwin.
Each reconstruction was compared against every ground-truth reference, and we evaluate resolution at an FSC of 0.5 since we have a reference structure and do not use half-maps.
Because we a priori do not know which cluster belongs to which structure, we exhaustively search for pairings which produce the lowest mean resolution.

\paragraph{Latent $k$-Means clustering.}
Clustering used scikit-learn’s KMeans with k-means++ initialization, 10 initializations, and a fixed random seed of 0. The solution with the lowest within-cluster sum of squared Euclidean distances was retained.
We use a convergence tolerance of $1e^{-4}$, and always set $k$ to the number of ground truth structures.

\paragraph{CryoDRGN-AI settings.} 
We use cryoDRGN-AI 4.3.0 and a latent dimension of eight.
The decoder output and loss are on Fourier downsampled $128^2$ images, but the encoders are always provided full resolution images.
This is done to make the cryoDRGN-AI training computationally tractable.
10,000 images are used to initialize the decoder with random poses, followed by three epochs of hierarchical pose search. 
Pose search retained eight candidate poses per image and used four refinement iterations. 
The translation-search grid used seven points per axis, and an extent parameter of 20 pixels. 
We use Adam with a learning rate of $1e^{-4}$ for the reconstruction network and encoder head, and $1e^{-2}$ for the conformation table when applicable.
These values were chosen from the recommended defaults.
Batch size was set to 16.
The decoder network uses a width of 256, three layers, and Gaussian positional encoding with dimension 64 and feature scale 0.5 in the Hartley domain. 

\paragraph{Probe Head Structures.} 
We use a dense spatial head on the two-dimensional grid of ViT patch tokens. 
It comprised a $1\times1$ convolution followed by three $3\times3$ convolutions with stride two, with output channel widths of 256, 384, 512, and 512. 
Each layer uses group norm with 32 groups, GELU activations, and a final tanh activation.
This is then projected down to an eight-dimensional latent space.

% \paragraph{Compute Resolution.} \nc{todo, should be where we talk about fsc optimal matching}

% \paragraph{Seeding (?)} \nc{todo, i know theres some seeding but have to check what}

\subsubsection{Hsp90 Details}

\paragraph{Image Generation.}
The dataset contains 100,000 simulated particle images, with 50,000 images from each of two conformations corresponding to CMA angles of 0$^\circ$ and 19$^\circ$ with CMB fixed at 0$^\circ$. 
Simulations used $384\times384$-pixel images at 1.5 \AA/pixel, a $480\times480$ padded simulation grid, deposited atomic B-factors, an accelerating voltage of 300 kV, spherical aberration of 2.7 mm, amplitude contrast of 0.1, and zero phase shift. 
Orientations are sampled uniformly over SO(3), and translations were sampled uniformly from -10 to +10 \AA. 
Mean defocus, astigmatism magnitude, and astigmatism angle were sampled uniformly over 10,000--15,000 \AA, 0--100 \AA, and 0--180$^\circ$, respectively. 
Gaussian white noise was added independently to each image with variance set to the clean-image variance divided by 0.01.

\begin{table}[htbp!]
      \centering
      \caption{
          Class-wise reconstruction resolution at FSC${}=0.5$ for Hsp90 at
          an SNR of 0.01, reported in \AA. Reference class 0 corresponds to the
          closed state, and reference class 1 corresponds to the open state. Lower is better.
          The corresponding ground-truth resolutions are 4.48\,\AA\
          (closed) and 4.50\,\AA\ (open).
      }
      \label{tab:hsp90-snr0p01-class-resolution}
      \small
      \setlength{\tabcolsep}{4pt}
      \begin{tabular}{lcccc}
          \toprule
          & \multicolumn{2}{c}{Epoch 1}
          & \multicolumn{2}{c}{Epoch 2} \\
          \cmidrule(lr){2-3}
          \cmidrule(lr){4-5}
          Method
          & Closed & Open
          & Closed & Open \\
          \midrule
          
         \rowcolor{paper} Oracle
          & 4.48 & 4.50
          & - & - \\

          \rowcolor{paper} Lookup Table
          & 8.71 & 9.12
          & 8.78 & 9.10 \\

          \rowcolor{paper} CNN Encoder
          & 7.27 & 6.43
          & 5.08 & 4.78 \\
          
          \rowcolor{paper} Cryo-IEF-B
          & 6.43 & 5.46
          & 7.42 & 6.08 \\

          \rowcolor{paper} CryoUNI-B
          & \textbf{5.37} & \textbf{5.39}
          & 4.78 & 4.97 \\

          \rowcolor{paper-deep}\model-B
          & 6.57 & 7.56
          & \textbf{4.53} & \textbf{4.55} \\

          \bottomrule
      \end{tabular}
\end{table}

\subsubsection{Tomotwin-8 Details}

\paragraph{Image Generation.}
The TomoTwin reconstruction dataset has 100,000 images distributed equally across eight atomic models, with 12,500 images per class. 
These 8 structures were selected by ordering the structures from CryoBench~\citep{jeon_cryobench_2024} according to amino acid count, and taking 8 equally spaced structures.
Images were generated using the same projection, orientation, translation, and noise procedures as HSP90, with a target SNR of 0.01. 
Simulations used $832^2$ box size at 0.5 \AA/pixel.
Microscope settings were 300 kV accelerating voltage, 2.7 mm spherical aberration, 0.1 amplitude contrast, and zero phase shift. 
Mean defocus and astigmatism magnitude were sampled uniformly over 5,000–21,000~\AA~and 0–1,000~\AA, respectively, with astigmatism angles uniform over 0–$180^\circ$.

% \begin{table}[!htbp]
%       \centering
%       \caption{
%           Class-wise reconstruction resolution at FSC${}=0.5$ over the first
%           three training epochs. Values are mean [min--max] across eight
%           reconstructed classes, reported in \AA. Lower is better.
%           The ground-truth reference resolution is 3.56\,\AA.
%         \nc{TODO: replace with fsc curves or actual resolution counts}
%       }
%       \label{tab:tomotwin-res-by-epoch}
%       \small
%       \setlength{\tabcolsep}{4pt}
%       \begin{tabular}{@{}lccc@{}}
%           \toprule
%           Method
%           & Epoch 1
%           & Epoch 2
%           & Epoch 3 \\
%           \midrule
%           CryoDRGN-AI
%           & 222.93 [222.89--222.99]
%           & 182.96 [115.64--223.02]
%           & 130.92 [3.85--222.51] \\

%           CryoDRGN-AI---CNN
%           & 29.93 [3.62--146.79]
%           & 10.04 [3.54--54.07]
%           & 3.76 [3.54--4.23] \\

%           Ours (frozen encoder + CNN)
%           & \textbf{3.74 [3.65--4.00]}
%           & \textbf{3.57 [3.53--3.62]}
%           & \textbf{3.56 [3.53--3.61]} \\
%           \bottomrule
%       \end{tabular}
%   \end{table}

% \input{appendices/pose-probes}
\clearpage
% \input{appendices/limitations}

%%%%%%%%%%%%%%%%%%%%%%%%%%%%%%%%%%%%%%%%%%%%%%%%%%%%%%%%%%%%

\newpage

\end{document}